\documentclass[pdflatex,sn-basic,iicol]{sn-jnl}

\usepackage{xcolor}
\usepackage{manyfoot}
\usepackage{textcomp}
\usepackage[title]{appendix}
\usepackage{amsmath}
\usepackage{amssymb}
\usepackage{amsthm}
\usepackage{mathrsfs}
\usepackage{graphicx}
\usepackage{multirow}
\usepackage{booktabs}
\usepackage{array}
\usepackage{url}

\theoremstyle{plain}

\theoremstyle{definition}

\begin{document}

\title[Losing the name before the box]{Losing the name before the box: measuring and repairing
what narrow fine-tuning costs a detector outside its deployment vocabulary}

\author[1]{\fnm{Trung Minh} \sur{Bui}}\email{minhtrung@keti.re.kr}
\author[1]{\fnm{Jongsul} \sur{Moon}}\email{moonjongsul@keti.re.kr}
\author[1]{\fnm{YoungOuk} \sur{Kim}}\email{kimyo@keti.re.kr}
\author[1]{\fnm{Jung-Hoon} \sur{Hwang}}\email{hwangjh@keti.re.kr}
\author*[1]{\fnm{Dongin} \sur{Shin}}\email{di\_shin@keti.re.kr}

\affil*[1]{\orgdiv{Intelligent Robotics Research Center}, \orgname{Korea Electronics Technology Institute}, \orgaddress{\city{Seongnam}, \postcode{13449}, \country{Republic of Korea}}}

\abstract{\unboldmath%
A detector pretrained on a broad corpus is fine-tuned on a narrow domain, its in-domain
accuracy improves, and it ships. We ask what happens meanwhile to its coverage of objects the vocabulary never names, which in obstacle detection and inspection carry the risk.
No in-domain test set holds an example of one.

We give a longitudinal protocol: one pretrained checkpoint against its own fine-tuned descendants. It tracks \emph{held-out top-$K$ proposal coverage} $C_\tau$: of categories pretraining covered and the vocabulary omits, the share of boxes a detector's top $K$ regions still cover. The quantity is the open-world proposal literature's; the longitudinal reading is not.

$C_\tau$ falls while in-domain accuracy rises, on four architectures and three domains,
by $5.12$ to $63.35$ points at IoU~$0.7$ on boxes above $1024\,\text{px}^2$. No in-domain number identifies the
fall, and neither does detection average precision, which charges a missed and a misnamed box alike. On the one architecture scoring both, adaptation costs $87\%$ of the AP against a fifth of
the coverage, and the naming goes first at all six depths of its freeze ladder, every run.

What breaks is structured: three architectures sharing no pretraining run agree on which categories lose coverage, and those a model never learned do not lose any. A repair follows
and needs no training: mixing a quarter of the pretrained \emph{state} back, normalisation
statistics included, raises coverage on every cell swept for at most $2.47$ points of in-domain accuracy. Seeing it costs one extra evaluation pass.
}


\maketitle
\clearpage

\section{Introduction}
\label{sec:intro}
\begin{figure*}[t]
\centering
\includegraphics[width=\linewidth]{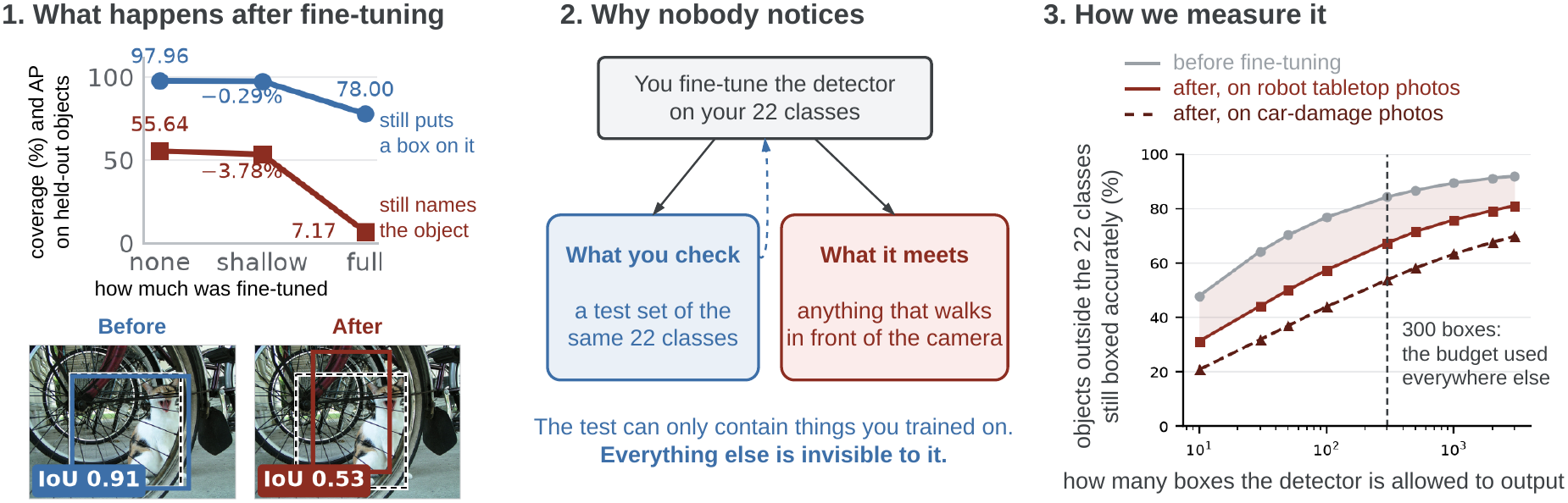}
\caption{\textbf{Fine-tune a detector on a narrow set of classes and it gets worse at finding everything else. The test that signs it off cannot show you.}
\textbf{1}~OWLv2, five seeds, on the $21{,}861$ COCO objects no deployment vocabulary names. One
line is the share of them it still puts a box on; the other is its detection AP on the same boxes.
Both run $0$ to $100$ but are not the same unit, so the two percentages give what each has lost
\emph{of its own starting value} (\S\ref{sec:namefirst}). Below, the same two checkpoints on one
image. IoU measures how much a predicted box overlaps the true one, so the photographs show the \emph{boxing}, the half that largely survives. The pair is the median of $586$ qualifying boxes, not an extreme one (\S\ref{sec:qualitative}).
\textbf{2}~The in-domain test set holds no such object, so it cannot report any of this.
\textbf{3}~What we count, at every budget. The gap is still open at $3000$ boxes, so the budget of
$300$ used elsewhere does not create it (\S\ref{sec:budget}).
Panels 1 and 2 are the problem and panel 3 the instrument; the repair is Figure~\ref{fig:wiseft}.}
\label{fig:teaser}
\end{figure*}

\begin{table*}[t]
\centering
\footnotesize
\setlength{\tabcolsep}{4pt}
\caption{\textbf{Everything this paper claims, and what each claim was tested on.} Ten claims in the four
groups the contributions are stated in; {$\bigstar$} marks the five that are new here and {$\bullet$} the
rest, which confirm or adapt a result the literature already carries (\S\ref{sec:related}).
\textbf{Every row's principal measurement rests on five or more independent seeds}; where a row
adds a supporting leg carrying fewer, the middle column or the section it names gives the count.
The middle column says how \emph{widely} each claim was tested. Four further claims, resting on less and argued only in the supplement, are in Table~\ref{tab:claimsfull}.}
\label{tab:claims}
\setlength{\tabcolsep}{4pt}
\begin{tabular}{p{7.6cm}p{5.9cm}l}
\toprule
claim & replicated over & where \\
\midrule
\multicolumn{3}{@{}l}{\textbf{What happens}} \\[1ex]
\addlinespace[0.25em]
{$\bullet$}\ A detector fine-tuned on a narrow vocabulary puts a box on fewer objects outside it, while its in-domain accuracy improves & 4 arch, 3 domains, 5 seeds, 2 further image sets, 2 checkpoints & \S\ref{sec:collapse} \\
\addlinespace[0.55em]
{$\bullet$}\ In-domain accuracy does not determine held-out coverage: across five seeds of one recipe it varies by $1.54$~mAP while coverage varies by $2.25$ points, and the most accurate seed is the least covered & 5 seeds within one recipe; a constructed cross-recipe pair differs by $11.37$ points, and a disjoint-support argument covers any in-domain functional & \S\ref{sec:invisibility} \\
\addlinespace[0.55em]
{$\bigstar$}\ \textbf{The name goes before the box.} After adaptation the model still puts a region on about three quarters of these objects and keeps an eighth of its AP on them, and already at a shallower freeze depth every seed has given up more of its naming than of its boxing & 1 arch, the only one scorable by name; 1 domain; all 6 rungs of its ladder, 3 or 5 seeds each but one, every run agreeing at both thresholds and every replicated rung clear of zero & \S\ref{sec:namefirst} \\
\midrule
\multicolumn{3}{@{}l}{\textbf{Which objects it happens to}} \\[1ex]
\addlinespace[0.25em]
{$\bigstar$}\ \textbf{What is damaged is what the checkpoint already had.} Categories the pretrained model never learned do not lose coverage; they gain it & 3 arch, 752 LVIS categories, 5 seeds; all six cells clear of zero, every seed sharing its cell's sign & \S\ref{sec:retention} \\
\addlinespace[0.55em]
{$\bigstar$}\ \textbf{Which objects are lost belongs to the object, not the model.} Three architectures that share no pretraining run agree on which categories lose coverage & 3 arch, 2 domains, 333 categories, 5 seeds each; support-floor sweep and cross-domain placebo & \S\ref{sec:crossarch} \\
\midrule
\multicolumn{3}{@{}l}{\textbf{How they are lost}} \\[1ex]
\addlinespace[0.25em]
{$\bigstar$}\ \textbf{Ranked out at the loose threshold, moved at the tight one.} The object is still proposed at IoU~0.5 and no longer ranked into the budget; at IoU~0.7 about half the loss is the box having moved, and which of the two dominates there differs by architecture & the crossing, 2 arch, 2 domains, 5 seeds; the enumeration beside it is 1 arch, 3 seeds & \S\ref{sec:proxyvalidation}, \S\ref{sec:decompose} \\
\addlinespace[0.55em]
{$\bullet$}\ Background suppression accounts for a minority of the drop & 2 arch, 2 domains, 5 seeds; 20 of 20 seed pairs move the same way at IoU~0.5, 19 of 20 at IoU~0.7 & \S\ref{sec:negatives} \\
\midrule
\multicolumn{3}{@{}l}{\textbf{What to do about it}} \\[1ex]
\addlinespace[0.25em]
{$\bullet$}\ Freeze depth is an architecture-dependent control, not a recipe & 4 arch, 7 ladders; every inverting rung at 5 seeds & \S\ref{sec:twoladders} \\
\addlinespace[0.55em]
{$\bullet$}\ Interpolating a quarter of the pretrained state recovers coverage on every cell swept, for at most $2.47$ points of in-domain accuracy and no training run & 3 arch, 2 domains, 5 seeds & \S\ref{sec:mitigations}, Figure~\ref{fig:wiseft} \\
\addlinespace[0.55em]
{$\bigstar$}\ \textbf{The repair works through the normalisation statistics, and moves them in the direction the literature on them argues against.} Re-estimating on deployment data, the standard move, is indistinguishable in domain and costs $2.2$ to $4.8$ points of coverage; moving them back toward pretraining recovers most of the gap & 1 arch, 2 domains, 5 seeds; the prescribed arm is negative on every seed of one domain and clear of zero at both thresholds & \S\ref{sec:mitigations}, Table~\ref{tab:bnrecal} \\
\bottomrule
\end{tabular}
\end{table*}

Open-vocabulary detectors are pretrained on broad corpora and then fine-tuned on the narrow vocabulary of a deployment domain~\citep{zhu2024survey}, a practice whose cost to open-world generalisation the detectors' own authors have reported~\citep{minderer2023scaling}. A typical instance pairs a checkpoint pretrained on the $80$ COCO categories with $247$ labelled images and $22$ classes. The in-domain metrics improve, so the model ships.
Meanwhile the detector has become worse at putting a box on objects the deployment vocabulary
never named, and the in-domain metrics cannot report it. The test set that certified the adaptation
contains no such object, so the loss lies outside its support.

\begin{table*}[t]
\centering
\small
\caption{\textbf{Four quantities this paper keeps apart}, because they are easy to confuse and only the
first is measured here. Table~\ref{tab:instruments} sets it against the instruments that might have
been used in its place.}
\label{tab:terms}
\begin{tabular}{>{\raggedright\arraybackslash}p{3.4cm}p{11.4cm}}
\toprule
term & what it denotes \\
\midrule
held-out top-$K$ proposal coverage &
\textbf{What we measure.} Of the objects whose class the deployment vocabulary never names, the
share that at least one of the detector's top $K$ boxes still lands on
(Equation~\ref{eq:coverage}). \emph{Held out} means held out of the \textbf{deployment
vocabulary}, not of pretraining: the pretrained model was trained on every one of these classes,
so this is what a checkpoint \emph{keeps}, not what it generalises to \\
\addlinespace
class-agnostic localisation &
\textbf{What it stands in for.} Putting a box on an object whose name the model was never given.
Not measured directly, and coverage can record a miss while this is intact \\
\addlinespace
open-world proposal recall &
\textbf{Where it comes from.} Average recall on categories held out of \emph{training}, an
established measurement~\citep{kim2022learning,konan2022extending}. We keep it and change its use:
from scoring a training recipe to reading one checkpoint before and after adaptation \\
\addlinespace
unseen-vocabulary recognition &
\textbf{A different thing.} Whether the model can still \emph{name} categories withheld from
adaptation. OWLv2 only, in mAP, never mixed into the coverage axis (\S\ref{sec:vocab}) \\
\bottomrule
\end{tabular}
\end{table*}

\textbf{Where this matters.} The objects that carry the risk are the ones outside the label set, which is the ordinary case in obstacle detection, inspection and monitoring. For a monitoring
system a lost proposal costs a missed record. For a safety-critical perception stack it is an
object the downstream planner never sees at all, and \S\ref{sec:limitations} says where that turns
this measurement into a release gate.

Coverage of $21{,}861$ COCO boxes whose categories the deployment vocabulary omits moves the
other way, on every one of four architectures (Table~\ref{tab:collapse}). OWLv2 loses the most,
from $97.96$ to $78.00$ at IoU~0.5 and from $90.06$ to $37.82$ at IoU~0.7, where every loss is
larger. Figure~\ref{fig:teaser} is the result in one picture.

\textbf{Held-out top-$K$ proposal coverage}, written $C_\tau$, is the fraction of ground-truth
boxes, in categories a deployment vocabulary does not name, for which at least one of a detector's
top-$K$ regions reaches IoU $\geq \tau$ (\S\ref{sec:objectness}). Every empirical claim below is
a claim about $C_\tau$, and Table~\ref{tab:terms} separates it from the three quantities it is
easiest to confuse it with.

Recall on categories a detector was not trained to name is an established
measurement~\citep{kim2022learning,konan2022extending,hosang2016what}, and it is the family
$C_\tau$ belongs to. We inherit that measurement and change its use. Open-world proposal work scores one training
recipe against another. We score one pretrained checkpoint against its own fine-tuned descendants,
across architectures, freeze depth, deployment domains and the interpolation path between the two
checkpoints. That path is the robust-fine-tuning
construction of \citet{wortsman2022robust}, used here to repair a capability rather than to raise
accuracy under distribution shift.

$C_\tau$ is not a measurement of \emph{class-agnostic localisation}, the general ability to put a
box on an object whose name the model was never given. Coverage at one budget and two thresholds
is one consequence of that ability, not the ability itself. A model can keep the ability and still
lose $C_\tau$, either by reordering its proposals or by moving one slightly across the threshold.
\S\ref{sec:proxyvalidation} measures how much of the fall each of those two accounts for.

Two adaptation recipes reaching the same in-domain accuracy, $18.16$ mAP on cardd-22, differ by
$11.37$ points of $C_\tau$. Within one recipe, five seeds span $2.25$ points, and the
highest-accuracy seed is the worst of them (\S\ref{sec:invisibility}). An in-domain test set holds no object the model has no
name for, so nothing computed on it separates such a pair. That holds whether the functional is
mAP, AP50, AR100, recall or calibration. This is a statement about what an in-domain
number \emph{determines}, not about correlation: the two may still correlate, and
\S\ref{sec:twoladders} shows that they do.

This sits inside catastrophic forgetting~\citep{mccloskey1989catastrophic} rather than beside it. It is one component of it, measured apart from the others. \citet{mai2024calibrated} show that much of what fine-tuning appears to destroy on other classes is ``the discrepant logit scales between the fine-tuning classes and the other classes'', which calibration undoes afterwards. That account cannot reach the quantity measured here: their remedy is a monotone rescaling of logits, and no monotone rescaling changes which $K$ candidates are largest, so it moves $C_\tau$ by exactly zero (\S\ref{sec:rescore}). The standard defences constrain the weights or replay old data, and detection has its own line of them. Both report preservation as detection AP, on base and novel splits or in aggregate (\S\ref{sec:related}); \citet{minderer2023scaling} report the open-world ability falling in proportion to fine-tuning duration. Detection AP charges both failures to one number: a box no longer put on the object, and a box put on it and named wrongly. It cannot say which occurred.

On the one architecture that can be scored both ways, the two separate widely and in an order. After fine-tuning OWLv2 still proposes about three quarters of those objects while its detection AP on them falls from $55.64$ to $7.17$, and the naming goes first at every depth of its freeze ladder (\S\ref{sec:namefirst}).

\paragraph{Contributions.} The measurement is the methodological one: held-out top-$K$ proposal coverage read \emph{longitudinally}, one pretrained checkpoint against its own fine-tuned descendants rather than one training recipe against another. It comes with the invariance
proposition of \S\ref{sec:app:invariance} that decides which freeze depths can move the number at
all. It costs one evaluation pass, needs no held-out labels beyond those a public benchmark already
carries, and is released as \texttt{covprobe}. What it establishes falls in four groups, and
Table~\ref{tab:claims} is the same ten statements with the evidence behind each.

\emph{What happens.} (1) $C_\tau$ falls while in-domain accuracy rises, on four architectures and
three deployment domains. (2) No functional of the in-domain test distribution identifies the fall,
and a constructed pair of checkpoints matched on in-domain accuracy differs widely on it. (3) The name goes before the box, and detection AP cannot report it.
On the one architecture where both are defined, adaptation costs $87\%$ of the AP against
a fifth of the coverage, and the ordering holds at all six depths of its freeze ladder, in every
run behind every depth, by $3.49$ points at the knee at IoU~$0.5$ (\S\ref{sec:namefirst}).

\emph{Which objects it happens to.} (4) What is damaged is what the checkpoint already had: over
$752$ LVIS categories split on pretraining exposure alone, the ones the model never learned do not
lose coverage at all, they gain it, while the ones it learned lose between $2.75$ and $9.31$
points. The two groups move in opposite directions on all three architectures. (5) Three
architectures sharing no pretraining run agree on which categories lose coverage, which makes the
spread across categories a property of the category rather than of any one checkpoint.

\emph{How they are lost.} (6) At IoU~$0.5$ the fall is mostly which candidates the model ranks into
its budget; at IoU~$0.7$ about half of it survives as a coordinate change that no reordering
recovers, and which of the two dominates there differs by architecture. (7) The account the
open-world literature would give, that adaptation pushes unmatched regions to background, is
measured on both sides and carries a minority of the drop.

\emph{What to do about it.} (8) Freeze depth is an architecture-dependent control rather than a
recipe, so we give \textbf{a measurement procedure} in its place: run the stages, discard those
whose in-domain accuracy says the fit failed, and ship from the Pareto front. (9) We give a repair that needs no training run: interpolate the pretrained \emph{state} back at $\alpha = 0.25$, a fraction a stated in-domain budget returns and which survives being chosen leave-one-cell-out. It recovers coverage on every architecture and domain swept for at most $2.47$ points of in-domain accuracy. (10) That repair works through the normalisation statistics, and moves them the way the literature on them argues against. Re-estimating on deployment data is the standing recommendation. On the one architecture where the two halves of the state were separated it is worse than leaving them fine-tuned on one domain, and recovers about a third of the gap on the other (\S\ref{sec:mitigations}).

\paragraph{The claims and their scope.} Table~\ref{tab:claims} states every claim, the evidence behind
it and the scope it is made at. Each row carries its own limits where the claim is made.

\texttt{covprobe} returns $C_\tau$ for a checkpoint, and an architecture enters it through one
function. Code and every measurement accompany this submission,
including the measurements a fault invalidated and the runs that replaced them, and will be
released publicly on publication. Rebuilding any coverage number reported here from the per-image counts behind it needs no GPU. \S\ref{sec:related} places the result against existing accounts of forgetting.
\S\ref{sec:objectness} defines the measurement and argues that no in-domain evaluation determines
it. \S\ref{sec:collapse} reports the coverage drop and \S\ref{sec:mechanism} asks what is
associated with it. \S\ref{sec:ladder} turns that into practice and \S\ref{sec:mitigations} prices
the repair, \S\ref{sec:vocab} extends it from
localisation to unseen vocabulary, and \S\ref{sec:limitations} states what the measurement does
not establish. Appendix~\ref{sec:appendix} opens with a map of its supporting measurements.

\section{Related work}
\label{sec:related}

Several literatures bear on this measurement, and each is taken in turn.
Table~\ref{tab:protocols} sets out where each one's protocol differs from ours; the paragraphs
below say what each measures and what its evidence leaves open.

\subsection{The measurement is inherited; the longitudinal use is not}
\label{sec:agnosticrecall}

Recall on categories a detector was never trained to name is established, and the quantity used
here belongs to that family. \citet{kim2022learning} report average recall at proposal budgets
from ten regions to a thousand on categories held out of training, which is
Equation~\ref{eq:coverage} at a different set of budgets. \citet{konan2022extending} name the
setting Open-World Proposals and score average recall on novel classes, including AR@300 on
LVIS~\citep{gupta2019lvis}. Average recall at a proposal budget is itself \citet{hosang2016what}'s.
Computed beside $C_\tau$ on our own population it agrees on sign and on the ordering of the cells. Its drop falls between the two thresholds' in all four (Appendix~\ref{sec:app:ark}), so neither of those turns on our choice of $\tau$. Its level is lower
throughout, which it must be, since it averages recall out to IoU~$0.95$. Open-world detection carries the same idea as U-Recall, counting unknown objects a
model localises without naming them~\citep{joseph2021towards}. \citet{zhu2024survey} survey the
wider open-vocabulary literature these measurements sit in, where the generalisation at stake is
to categories outside a model's training vocabulary.

\textbf{Decomposing AP asks which failure occurred, of a different population.}
\citet{hoiem2012diagnosing} sort a detector's errors into named types and \citet{bolya2020tide}
weight each by the average precision it costs, separating a region localised well and misnamed from
one missed altogether. That accounting runs over the detections a model emits; ours runs over the
regions it proposes (Table~\ref{tab:instruments}).

In each of those evaluations the object under study is a \emph{design} and the recall figure is
the score that design earns. Ours is a checkpoint that already has the recall, and although the
instrument is theirs, the question is what a fine-tune on a few hundred narrow-domain images does
to it.
The categories are held out of the \emph{deployment vocabulary} while remaining in pretraining, so
the quantity is retention of something the checkpoint demonstrably had. The comparison is also longitudinal within one checkpoint rather than across designs.

\subsection{Mechanisms proposed for the drop, and what each protocol can see}

Prior work offers two accounts of \emph{why} an adapted detector stops proposing things. A
classification objective may teach the proposal mechanism to suppress what it was not asked to
find, or forgetting may concentrate in particular components. The first account is still being
built on: \citet{kormushev2026mitigating} attribute lost masks in open-vocabulary panoptic
segmentation to ``objectness heads trained on closed vocabularies'' and adjust the objectness
probability to recover them. \S\ref{sec:negatives} measures how much of a detector's coverage
loss that pathway can carry. Incremental
detection asks the neighbouring question of what a detector loses when its label set
\emph{grows}~\citep{shmelkov2017incremental,wu2025demystifying}, where ours narrows. Which of
these survives contact with our data is tested in \S\ref{sec:mechanism}, and the literature each belongs to is set out in \S\ref{sec:app:related}.
\label{sec:oln}

\subsection{What to move, and what to leave alone}

Where to intervene has been studied without reference to proposal coverage:
\citet{lee2023surgical} show the layers worth tuning depend on the kind of shift, and
\citet{ruis2025textual} adapt an open-vocabulary detector on the text side with the backbone
frozen. Weight interpolation is established as a
patch~\citep{wortsman2022robust,ilharco2022patching}. The repairs compared in
\S\ref{sec:ladder} are drawn from those families and priced against each other on one axis
(\S\ref{sec:app:alternatives}).

A separate line treats a network's normalisation statistics as a domain-carrying object that can
be acted on by itself. \citet{li2016adabn} adapt by recomputing them on target data with nothing
trained; \citet{schneider2020improving} report the same swap improving corruption robustness across
$25$ models; \citet{nado2020prediction} do it at prediction time; \citet{wang2021tent} optimise
through them at test time. That literature prescribes moving the statistics \emph{toward the deployment domain}, and is judged on accuracy there. \S\ref{sec:mitigations} runs that arm and finds it does what the literature promises while costing held-out coverage. In-domain accuracy is indistinguishable from the arm we ship; held-out coverage falls. These are two objectives rather than a contradiction, and the one that serves the first is the wrong direction for the second. Two later methods stop short of replacing the source
statistics: \citet{lim2023ttn} interpolate source and test statistics with a learned per-layer
weight, and \citet{su2024tema} accumulate them across test batches. Both keep some of the source
estimate, which is the direction our result points, although both fit or accumulate over the
statistics alone where we interpolate the whole state with one scalar. Neither is tested here. \citet{nado2020prediction} already flag the interaction, reporting mixed results for the method
alongside pre-training, and \citet{neyshabur2020what} separate low-level input statistics from
feature reuse as distinct carriers of what transfer transfers. Nor is it new that the two halves of
the state need separate treatment. \citet{li2024merging} merge domain-adapted models by handling
``model parameters and model buffers'' as two problems, where linear merging suffices for the
first although the second needs a fitted prior. What \S\ref{sec:mitigations} adds is a price for the split on one architecture.

Averaging as a repair is also studied outside distribution shift. \citet{kleiman2025soup} merge a
training model with its own earlier checkpoints and beat Task Arithmetic, TIES, WiSE-FT, L2 and
EWC on sequential tasks, a comparison set that overlaps Table~\ref{tab:alternatives}. \S\ref{sec:alternatives} finds averaging along the trajectory
recovers nothing here. The two are consistent: their averaging spans tasks, ours would span
iterates already inside one fine-tuned basin. Cross-domain few-shot detection reaches the same concern from
the accuracy side, with \citet{yu2026closer} retaining pretrained weights through fine-tuning and
scoring the result on further detection benchmarks rather than on proposal coverage.

\subsection{What this paper adds}

Table~\ref{tab:protocols} places this work as a combination of two existing protocols rather than a
new instrument. The open-world proposal literature separates proposing from naming and measures it
carefully, on designs trained from scratch with the evaluated categories deliberately withheld. The
forgetting and robust-fine-tuning literatures follow one pretrained checkpoint through adaptation,
which is our comparison, and score it with detection AP or classification accuracy. Those aggregates
charge a missed box and a wrong name to one number, and \S\ref{sec:intro} measures how far apart
the two come on the one architecture that defines both. Reading a class-agnostic, pre-NMS coverage
longitudinally, on categories the checkpoint was pretrained on, is what separates them. We show alongside it that under an
unrestricted hypothesis class the quantity is not identifiable from any in-domain evaluation
functional, and give the measured pair of checkpoints that carries the point empirically
(\S\ref{sec:invisibility}).

That leaves the question of whether this is forgetting measured with a different metric.
\citet{wu2025demystifying} give the sharpest prior answer, localising forgetting in two-stage
incremental detection to the RoI-head classifier ``rather than degradation in RPN's recall
capability''. That is the prior result closest to contradicting this one. The two protocols
differ in what each holds fixed. Their splits draw every stage from one image distribution and
change only the label set, so their RPN never meets a distribution it was not pretrained for. Here the adaptation images move and the evaluation stays on the original distribution.
Neither protocol can name the fragile component alone, and together they say the locus depends on
which shift is imposed (\S\ref{sec:app:related}). Two components move here, and
they are measured on different populations of architecture. Naming and boxing separate on the one
architecture that scores both (\S\ref{sec:namefirst}). Inside coverage itself, on the three architectures whose candidates carry an
index, the fall at IoU~$0.5$ is predominantly which candidates are selected rather than how they
are named, and it appears in one-stage anchor and query-based detectors with no RoI head at all.
The two readings are not in tension: the first compares two metrics on the same boxes, the second
decomposes one of them. \citet{robinson2026rfdetr}, whose detector is one of the four measured
here, reach the naming half from the other side. Prompting a fine-tuned Grounding DINO with real
class names rather than class indices buys almost nothing across a hundred narrow domains, which
they read as naive end-to-end fine-tuning ``mitigat[ing] the benefits of open-vocabulary
pre-training''. Theirs is presence against absence on one comparison; what is measured here is how
much of the naming goes against how much of the boxing.
\citet{zhang2026orthogonal} define a class-agnostic object recall for the reason $C_\tau$ is
defined here, that ``unlike mAP, oRecall directly quantifies the model's ability to localize
foreground objects independent of category-level classification''. They attribute its fall under
full and backbone tuning to ``corrupting the previously learned objectness scoring and feature statistics'', the two components \S\ref{sec:proxyvalidation} and \S\ref{sec:mitigations}
separate and price. Their label space is fixed while the image domain moves; ours is the deployment
vocabulary narrowing, so their held-out categories were in training and ours are held out of it.
The damage also has a structure a forgetting account does not predict. Architectures that share no
pretraining run agree on which categories lose coverage, and a category's fate follows its prior
coverage rather than its name (\S\ref{sec:crossarch}, \S\ref{sec:retention}). Two
category-independent rates then reproduce the aggregate on a benchmark they were never fitted to
(\S\ref{sec:flipmodel}). A coverage drop depends on the
benchmark it was read on as much as on the model. A rate pair describes the adaptation itself, so
two studies measuring the same damage on benchmarks of different baseline coverage will report
different effect sizes without either being wrong.

\section{Measurement protocol}
\label{sec:objectness}

\subsection{Held-out top-$K$ proposal coverage}

Let $\mathcal{R}_K(M, I)$ be the $K$ regions detector $M$ retains on image $I$ under its own
selection rule. That rule is an objectness ranking where the architecture has one, the whole query
set where it does not, and a text-conditioned ranking for Grounding DINO; Table~\ref{tab:candidates}
gives each. Let $\mathcal{B}$ be a set of (image, ground-truth box) pairs, and write
\begin{equation}
\label{eq:bestiou}
\mathrm{IoU}^\star(M, I, b) \;=\; \max_{r \,\in\, \mathcal{R}_K(M,\,I)} \mathrm{IoU}(r, b)
\end{equation}
for the overlap the best of those $K$ regions reaches on box $b$. This is the quantity
\S\ref{sec:decompose} takes apart, since a region can lose it either by falling out of
$\mathcal{R}_K$ or by moving. For an overlap threshold $\tau$,
\begin{equation}
\label{eq:coverage}
C_\tau(M, \mathcal{B}) \;=\; \frac{1}{|\mathcal{B}|}
\sum_{(I,\,b)\,\in\,\mathcal{B}}
\mathbf{1}\!\left[\,\mathrm{IoU}^\star(M, I, b) \geq \tau\,\right].
\end{equation}
This is recall without classification: labels and confidences are discarded, and the question is
whether the top-$K$ proposal set still covers the object, never whether the model can name it.

\begin{table}[t]
\centering
\footnotesize
\setlength{\tabcolsep}{3pt}
\caption{\textbf{Why the standard instruments cannot report this loss.} Everything above the rule is read
from the detections a model finally emits, so it can speak only about categories the model still has a name for. A narrow fine-tune leaves it $22$. \citet{hoiem2012diagnosing} and \citet{bolya2020tide} come closest (\S\ref{sec:related}), but both take their bins after ranking and non-maximum suppression, so a region the
model still produces but no longer ranks into its output counts as \emph{missed} there and as
present here (\S\ref{sec:proxyvalidation}). \textbf{The last two rows read the same quantity of
different categories}: open-world proposal recall scores what a recipe never learned, $C_\tau$ what
this checkpoint knew and was not asked about, its top $K$ regions taken by the architecture's own
selection rule.}
\label{tab:instruments}
\begin{tabular}{@{}>{\raggedright\arraybackslash}p{2.05cm}>{\centering\arraybackslash}p{0.95cm}>{\centering\arraybackslash}p{1.45cm}>{\raggedright\arraybackslash}p{2.15cm}@{}}
\toprule
& needs & read before the & \\
instrument & a name & class score & categories it scores \\
\midrule
\multicolumn{4}{@{}l}{\emph{read from the detections a model emits}} \\
\addlinespace[0.25em]
detection AP & yes & no & the ones it names \\
\addlinespace[0.4em]
base/novel split AP~\citep{wu2025demystifying} & yes & no & the ones it names \\
\addlinespace[0.4em]
TIDE, Hoiem & yes & no & the ones it names \\
\midrule
\multicolumn{4}{@{}l}{\emph{read from the regions it proposes}} \\
\addlinespace[0.25em]
open-world proposal recall & no & yes & held out of \textbf{training} \\
\addlinespace[0.4em]
$C_\tau$, this paper & no & yes & held out of the \textbf{vocabulary}, present in pretraining \\
\bottomrule
\end{tabular}
\end{table}

\subsection{In-domain evaluation is not sufficient to identify held-out coverage}
\label{sec:invisibility}

Reaching $18.16$ mAP on cardd-22 with YOLOX two different ways covers $61.13$ and $72.50$ points
of the held-out group at IoU~$0.7$. The first is a checkpoint anchored to the pretrained weights during
training; the second is the interpolation curve of \S\ref{sec:mitigations} read at that same in-domain
accuracy. That is the matched-accuracy comparison used throughout \S\ref{sec:ladder}, although no second training run stands behind it. No in-domain test set separates the two, and eleven points of
the quantity at issue lie between them.

Within a single recipe the gap is small: five seeds of the plain cardd-22 fine-tune span $1.54$
mAP and $2.25$ points of coverage, and the highest-accuracy seed has the \emph{lowest} coverage of
the five. Across recipes it is large, and $11.37$ points is the widest such pair we found, though a
typical one is smaller. At either scale the in-domain number does not determine the held-out one.

\textbf{The seed sweep.} The sweep is an underspecification diagnostic, and the name is \citeauthor{damour2022underspecification}'s. A pipeline is underspecified when it ``can return
many predictors with equivalently strong held-out performance in the training domain'' which
``behave very differently in deployment domains''~\citep{damour2022underspecification}. Varying
only the seed, holding in-domain accuracy nearly fixed and reading a property the pipeline never
constrained is their diagnostic, and $C_\tau$ is the property. The two claims are worth stating separately. The pipeline is underspecified with respect to
$C_\tau$, which is an empirical statement about these five checkpoints. No in-domain functional
identifies $C_\tau$, which is an idealised one and needs the assumption of the next paragraph.

In-domain accuracy still carries information. \S\ref{sec:twoladders} fits a regression in
which adding in-domain mAP raises $R^2$ from $0.05$ to $0.80$ on cardd-22, with a sign saying that
fitting the deployment domain better is protective. That regression is fitted across freeze depths
whose coverage was already measured, so it describes runs already scored. It does not let a
practitioner read the coverage of a checkpoint in hand off that checkpoint's in-domain number, and
it is that inference the measurement rules out.

\textbf{In-domain mAP as a selection rule.} The number is misleading on one architecture and uninformative on the other. The identifiability argument says prediction is impossible in
principle; the useful question is how well it works in practice. Take the decision directly:
several candidate adaptations of one detector for one domain, in-domain mAP measurable on all of
them, and the aim of keeping the most held-out coverage. The sweep is $110$ individual runs: every rung and seed of the YOLOX and RF-DETR~\citep{robinson2026rfdetr} ladders on both
domains. On YOLOX in-domain mAP is strongly \emph{negatively} rank-correlated with coverage,
$-0.918$ and $-0.840$ at IoU~$0.5$. Picking the higher-mAP of two candidates there returns the
higher-coverage one in $12$ to $21\%$ of pairs. On RF-DETR
it is near zero, $-0.163$ on tabletop-22 and $+0.015$ on cardd-22, the one cell where selecting on
mAP beats chance at all, and it beats it by $5$ points. Three of the four cells sit below chance at
both thresholds, and none is a cell where the in-domain number is a good selector.

Those two YOLOX cells run against a regularity that usually holds. \citet{miller2021accuracy}
report out-of-distribution performance ``strongly correlated with in-distribution performance for a
wide range of models and distribution shifts'', across architectures, hyperparameters and training
duration. Their axis is accuracy on both sides; ours is accuracy against coverage. Their model families
vary, where our runs vary only freeze depth and seed within one checkpoint. We test neither
difference. The negative sign is worth
reporting as a place the regularity does not reach rather than as a refutation of it.

That sits alongside the protective coefficient above without contradicting it, and the difference is
which variable is held fixed. Depth raises in-domain accuracy and lowers coverage, so mAP is
protective at fixed depth and anti-correlated with coverage when depth is free to move, which is
the position a practitioner comparing candidate rungs is in. Pairs within a cell share runs. The shares therefore carry no binomial interval, and the
replicating unit is the cell, of which there are four. On YOLOX the negative association restates the depth result of \S\ref{sec:twoladders} in the
units of a decision, adding no evidence to it. What it adds is that the restatement fails on
RF-DETR, whose ladder has no depth ordering to restate.

\subsection{Setup}
\label{sec:setup}

Coverage is read on COCO~\citep{lin2014coco} \texttt{val2017} at boxes of at least $1024\,\text{px}^2$ with no crowd
annotations, \textbf{$4877$ images and $24{,}086$ boxes}, of which $21{,}861$ lie outside every
deployment vocabulary and $2{,}225$ inside it as a control. Four architectures are read through
their top $K = 300$ regions pre-NMS: Faster~R-CNN~\citep{ren2015faster}, OWLv2~\citep{minderer2023scaling}
and YOLOX~\citep{ge2021yolox} rank by a native class-agnostic score, and RF-DETR~\citep{robinson2026rfdetr},
which has none, contributes its whole query set. Grounding DINO~\citep{liu2024grounding} is
reported separately (Appendix~\ref{sec:app:gdinodetail}). Deployment domains are
tabletop-22, cardd-22~\citep{cardd22hub}, not the CarDD benchmark, and bccd-3~\citep{bccd2017},
each $247$ images split $185/62$. Every measurement script asserts the evaluation set's size at run time. Recipes, score definitions, the budget sweep and the AR@K cross-check are in \S\ref{sec:app:budget} and \S\ref{sec:app:ark}.

\paragraph{What was run, and where the standing objections are answered.}
Table~\ref{tab:experiments} maps the paper's twenty-odd experiments onto the eight questions they
answer. Six objections recur, and each has a section rather
than a sentence: whether a fixed $K$ is fair across architectures emitting different numbers of
candidates (\S\ref{sec:budget}, \S\ref{sec:proxyvalidation}, \S\ref{sec:collapse}); whether the
open-world literature's own metric agrees, which it does (\S\ref{sec:aratk}); how the
interpolation fraction is chosen without held-out categories (\S\ref{sec:mitigations},
Appendix~\ref{sec:app:alphabudget}); whether the loss can be repaired at inference instead, which
it partly can (\S\ref{sec:rescore}); where the descriptive baseline fails, which is across box
scales (\S\ref{sec:retention}, Appendix~\ref{sec:app:bandlaw}); and whether this is catastrophic
forgetting under another name (\S\ref{sec:related}).

\textbf{Not every rung can move the number.} $C_\tau$ reads only the coordinates the ranking
score and the geometry read, so a rung that trains parameters outside that set cannot change coverage. Such a cell measures
$-0.00$ by construction, and finds nothing about whether adaptation is free. Which rungs those are and the proposition that fixes it are in Appendix~\ref{sec:app:invariance}, together with the harness fault it caught: a `frozen' ladder that was adapting every BatchNorm buffer in the backbone. What was fixed before measuring and what was searched is in
\S\ref{sec:app:prespec}, together with how to read every interval reported here
(\S\ref{sec:app:intervals}).

\textbf{The validity control.} A fine-tune that damages coverage because it failed at its own task
looks exactly like the finding. The criterion fixed before training is a strict inequality with
no margin: a run passes when its in-domain mAP on the held-out deployment split exceeds the
pretrained model's. Faster~R-CNN, RF-DETR and YOLOX replace their classification head, so for them
it reduces to $\text{mAP} > 0$. Five RF-DETR and YOLOX cardd-22 cells pass it at $0.16$ to
$8.13$~mAP and are set aside by a judgement made after seeing the data, that a rung below
$10$~mAP is not a working detector; the all-attempts estimand, which keeps them, therefore leads
throughout \S\ref{sec:ladder}, and every aggregate is reported both ways
(Appendix~\ref{sec:app:itt}). Every \texttt{full}-rung cell passes both. The pre-specified
criterion fails one cell of the \texttt{ignore} ablation (\S\ref{sec:ignorearm}) and all of
voc-20, by $18.90$~mAP (\S\ref{sec:limitations}).

\subsection{Which tests are confirmatory, and which are not}
\label{sec:teststats}

Seven hypotheses had their quantity, direction and test fixed before the measurement: that $C_\tau$
falls under full fine-tuning; that depth orders the drop within a ladder; that displacement is
associated with threshold crossing; that its constant is $2$; that interpolation is non-dominated
at matched accuracy; and, as predicted nulls that come back null, that vocabulary membership adds
predictive value and that the negative-label pathway carries the drop. Three fail: depth orders
the drop on YOLOX and inverts it on RF-DETR (\S\ref{sec:twoladders}), the constant moves with the
fitting window (\S\ref{sec:displacement}), and L2-SP beats interpolation on Faster~R-CNN
(\S\ref{sec:anchoring}). Everything else is exploratory, including which category property
predicts the residual, the matcher and fitting window, and $\alpha = 0.25$, and is reported with
the weak effects it earns, never as a finding. Where several tests share a family we apply Holm; where an effect size and its interval
carry the claim we report those and quote no $p$-value. Appendix~\ref{sec:app:stats} gives each formal test its class, correction and outcome.

\subsection{Use of large language models}

The authors set the research question, the quantity to measure and the experimental design, decided every claim and the scope it is stated at, and ran every training run and evaluation reported here on their own infrastructure. No result in this paper was produced by a language model.

A large language model was used as an assistive tool. It drafted text, and wrote most of the analysis and auditing code released with the submission, including the two scripts that re-derive the manuscript's numbers from the per-run files and fail on any disagreement. Those two scripts therefore share an author with the manuscript and are not an independent check: what they rule out is a number drifting from the file it came from, and every table they cover was also re-derived by hand at least once.

The authors are responsible for all of it, and specifically for every number and every citation.

\section{The coverage drop across architectures and domains}
\label{sec:collapse}

\begin{table*}[t]
\centering
\footnotesize
\setlength{\tabcolsep}{2.6pt}
\caption{\textbf{The share of objects each detector still puts a box on, before and after fine-tuning}, using its own top $300$ regions and no class label. \emph{novel} is the
$21{,}861$ COCO boxes no deployment vocabulary names, \emph{control} the $2{,}225$ that tabletop-22
does, so \emph{control} is a naming contrast on the tabletop-22 rows only and $\dagger$ marks every
cell where it is not. The last column drops the $1024\,\text{px}^2$ size floor
(Table~\ref{tab:sizebands}). \textbf{Each $\pm$ is one standard deviation over five seeds};
a cell without one is a single run, and \texttt{---} means not measured, or under \emph{trainable}
that the row trains nothing. Seeds spread widest on OWLv2's fine-tuned row, where the $95\%$
interval is half the drop it reports. Grounding DINO: Table~\ref{tab:gdino}.}
\label{tab:collapse}
\begin{tabular}{llrrrrrr}
\toprule
& & & \multicolumn{2}{c}{novel} & \multicolumn{2}{c}{control} & novel, \\
\cmidrule(lr){4-5}\cmidrule(lr){6-7}
architecture & condition & trainable & IoU 0.5 & IoU 0.7 & IoU 0.5 & IoU 0.7 & unfilt.\ 0.5 \\
\midrule
\multirow{4}{*}{Faster R-CNN}
 & pretrained (COCO) & --- & 95.10 & 84.26 & 94.11 & 84.00 & 88.54 \\
 & tabletop-22, +RPN & 14.6M & 91.81 {\scriptsize$\pm$0.23} & 77.32 {\scriptsize$\pm$0.11} & 95.16 {\scriptsize$\pm$0.11} & 84.30 {\scriptsize$\pm$0.20} & --- \\
 & tabletop-22, all & 41.2M & \textbf{85.14 {\scriptsize$\pm$1.00}} & \textbf{66.58 {\scriptsize$\pm$1.44}} & 85.63 {\scriptsize$\pm$1.11} & 71.59 {\scriptsize$\pm$1.16} & \textbf{74.96} \\
 & cardd-22, all & 41.2M & \textbf{78.71 {\scriptsize$\pm$0.51}} & \textbf{54.17 {\scriptsize$\pm$0.99}} & 75.83$^{\dagger}$ {\scriptsize$\pm$1.80} & 57.82$^{\dagger}$ {\scriptsize$\pm$1.64} & \textbf{62.78} \\
\midrule
\multirow{3}{*}{OWLv2}
 & pretrained & --- & 97.96 & 90.06 & 97.84 & 91.19 & 92.99 \\
 & fine-tuned, 2 vision layers & 16.9M & 97.68 {\scriptsize$\pm$0.09} & 88.80 {\scriptsize$\pm$0.70} & 97.54 {\scriptsize$\pm$0.29} & 89.90 {\scriptsize$\pm$0.82} & --- \\
 & fine-tuned, all & 155.0M & \textbf{78.00 {\scriptsize$\pm$8.91}} & \textbf{37.82 {\scriptsize$\pm$6.26}} & 84.99 {\scriptsize$\pm$5.38} & 48.99 {\scriptsize$\pm$5.73} & \textbf{74.78} \\
\midrule
\multirow{3}{*}{RF-DETR}
 & pretrained (COCO) & --- & 98.46 & 92.35 & 97.80 & 91.64 & 92.62 \\
 & tabletop-22, all & 31.9M & \textbf{97.17 {\scriptsize$\pm$0.31}} & \textbf{87.23 {\scriptsize$\pm$0.81}} & 96.31 {\scriptsize$\pm$0.11} & 87.63 {\scriptsize$\pm$0.68} & \textbf{88.93} \\
 & cardd-22, all & 31.9M & \textbf{92.60 {\scriptsize$\pm$0.66}} & \textbf{80.35 {\scriptsize$\pm$1.20}} & 84.35$^{\dagger}$ {\scriptsize$\pm$0.39} & 73.99$^{\dagger}$ {\scriptsize$\pm$0.48} & \textbf{84.21} \\
\midrule
\multirow{3}{*}{YOLOX}
 & pretrained (COCO) & --- & 92.09 & 84.16 & 86.70 & 80.18 & 82.66 \\
 & tabletop-22, all & 8.9M & \textbf{82.34 {\scriptsize$\pm$1.18}} & \textbf{71.29 {\scriptsize$\pm$1.23}} & 79.90 {\scriptsize$\pm$1.21} & 70.08 {\scriptsize$\pm$1.33} & \textbf{69.13} \\
 & cardd-22, all & 8.9M & \textbf{74.13 {\scriptsize$\pm$0.62}} & \textbf{60.37 {\scriptsize$\pm$0.90}} & 71.05$^{\dagger}$ {\scriptsize$\pm$1.19} & 60.12$^{\dagger}$ {\scriptsize$\pm$1.34} & \textbf{62.78} \\
\bottomrule
\end{tabular}
\end{table*}

Table~\ref{tab:collapse} carries the study's central measurement. All four architectures were trained and measured in one harness on the same
two $247$-image, $22$-class splits, and every fine-tuned row satisfies the validity control of
\S\ref{sec:control}. Grounding DINO fails in a way none of these four does, and
Appendix~\ref{sec:app:gdinodetail} takes it up separately.

\paragraph{The pretrained rows.} Before any adaptation all four land high on the novel group
(Table~\ref{tab:collapse}), although they differ in how many regions they emit and in what plays
the part of an objectness score: three read a \emph{native} class-agnostic score. RF-DETR has no objectness head at all, so the measurement takes its whole query set and consults no score
(\S\ref{sec:objectness}). No part of the protocol was tuned per architecture, and
the ordering among them does not follow which of them has a purpose-built objectness
score, so what the protocol is reading is a property of the models themselves.

Under a full fine-tune all four lose coverage on the novel group. Grounding DINO does too, although on a measurement that resists comparison with theirs (Appendix~\ref{sec:app:gdinodetail}). At IoU~0.5 the drop is
$9.96$ points for Faster~R-CNN on tabletop-22 and
$16.39$ on cardd-22, $19.96$ for OWLv2, $1.29$ and $5.86$ for RF-DETR, $9.74$ and $17.95$ for YOLOX. Every one grows at IoU~0.7, from $5.12$ for RF-DETR on tabletop-22 to $52.24$ for OWLv2. A fine-tuned model still puts a box near a novel object, but no longer around it, and the effect is larger than seed variation. Across the twelve RF-DETR rungs of
\S\ref{sec:twoladders} every cell rejects
no-change after Holm correction within that architecture, and so does every one of YOLOX's ten but
\texttt{preds} on tabletop-22 at IoU~$0.5$, which pins the candidate set to the pretrained model's
and measures $-0.00$. That rung leaves box regression free, so \S\ref{sec:invariance} does not
require it to read zero, and on cardd-22 it does not (\S\ref{sec:displacement}).

\subsection{The name goes before the box}
\label{sec:namefirst}

Detection AP charges two failures to one number: a detector that no longer puts a box on an object,
and one that boxes it and names it wrongly. No decomposition of it recovers them on these categories
(Table~\ref{tab:instruments}). OWLv2 separates them, because it keeps its vocabulary
in text and the same checkpoint is therefore scorable by name on the categories the deployment
vocabulary omits. The other three cannot: each replaces its classification head for the deployment
vocabulary and has no AP left on the remaining COCO categories.

\textbf{At full depth the two separate widely.} On the $21{,}861$ held-out boxes, fine-tuning takes
detection AP from $55.64$ to $7.17$ while coverage at IoU~$0.5$ falls from $97.96$ to $78.00$: the
adapted model still proposes about three quarters of these objects and keeps an eighth of its AP on them.
A detection benchmark re-run on it reports the $87\%$ fall and attributes none of it. Every seed
agrees on the direction and the order of magnitude, giving up $84.6$ to $89.9\%$ of its AP against
$9.5$ to $33.9\%$ of its coverage.

\textbf{And the naming goes first, at every depth of the ladder.} Every rung of
\S\ref{sec:twoladders} carries both quantities, so the comparison is paired within each checkpoint:
subtract the share of coverage it has lost from the share of AP it has lost. That
difference is positive at all six depths and in every seed of every one, and at each of the five
rungs carrying more than one run its interval clears zero at both thresholds. It is smallest at
\texttt{heads + objectness}, $+0.62$ points; $+3.49$ at the knee, with
$[2.87, 4.11]$ at IoU~$0.5$ and $+2.38$ with $[1.10, 3.66]$ at IoU~$0.7$; and $+66.72$ with
$[56.57, 76.87]$ at full depth. It is not monotone in depth: the shallowest rung of all, training
the heads alone, gives $+7.91$. What separates those two rungs is the naming side, $7.93\%$ of the
AP lost with the heads alone against $0.61\%$ one rung deeper, and in-domain mAP is non-monotone
across the same pair (Table~\ref{tab:owlv2ladder}). We have no account of it; the ordering is
positive at both. Neither a ratio of the two relative
losses nor a reading of the two intervals side by side would carry this, for reasons given with the
measurement in Appendix~\ref{sec:app:baseap}.

\subsection{The image partition, which the seed bars do not carry}
\label{sec:partition}

Error bars everywhere here are over model seeds at one image partition. A crossed design over four
further partitions puts the partition's own contribution well below the seed's, sd~$0.11$ at
IoU~$0.5$ and $0.09$ at $0.7$ against seed sd $0.95$ and $1.27$ on the same design, and an $F$-test
cannot distinguish the partition effect from zero at this replication
(\S\ref{sec:app:splitseed}).

\subsection{What the drop looks like on an image}
\label{sec:qualitative}

The selection rule was fixed before looking: a novel category from a pre-specified list, area at
least $4\%$ of the image, pretrained $\mathrm{IoU} \geq 0.85$ and fine-tuned
$\mathrm{IoU} \leq 0.60$. Over the first $1200$ \texttt{val2017} images it yields $19$ candidates
out of $2114$ eligible boxes on Faster~R-CNN. The rate matters more than the pictures, and
it is a property of the architecture rather than of the rule: the same $2114$ boxes and the same thresholds give $586$ candidates on OWLv2. That is the architecture that loses the most coverage in this study. On Faster~R-CNN the dramatic case is roughly one box in a hundred; on OWLv2 it is better
than one in four, which is why Figure~\ref{fig:teaser} shows the median of that pool rather than
its top. The examples themselves, the median of the selected pool and
three boxes drawn at random are in Appendix~\ref{sec:app:qualitative}, where they cannot be
mistaken for evidence about prevalence.

\subsection{Which objects the headline is about}
\label{sec:sizebands}

\textbf{Smaller objects lose more on four of the seven architecture-domain cells}
(Table~\ref{tab:sizebands}), which is what the geometry of
\S\ref{sec:displacement} predicts: a nudge that does not scale with the box carries a small one
across the overlap threshold and leaves a large one above it. Three cells do not, and on
those the conditional-mean baseline of \S\ref{sec:fliprates} shows through: a band that starts
low has little left to lose. YOLOX supplies two of the three: its smallest band starts at
$40.1\%$ pretrained coverage, the lowest starting point of any cell, and is not its worst band on
either domain. OWLv2's largest band is the third, losing more than the band below it from
starting points $1.5$ points apart.
Neither account orders the bands on its own: extrapolated to bands, the baseline
predicts loss \emph{growing} with the coverage a band starts from, which is the reverse of the
dominant trend and wrong in sign on the smallest band (Appendix~\ref{sec:app:bandlaw}).

RF-DETR's small figure on this population is also what the aggregate law of \S\ref{sec:fliprates} admits, and the cap that law implies belongs to the domain its rates were fitted on rather than to the architecture (Appendix~\ref{sec:app:cap}, \S\ref{sec:app:bandlaw}).

\textbf{The loss also concentrates by image, not only by box size.} $C_\tau$ averages over boxes,
so an image carrying many novel objects counts more than one carrying a single novel object.
Weighted per image instead, YOLOX's full fine-tunes lose less, $6.50$ points against $9.74$ on
tabletop-22 and $12.62$ against $17.95$ on cardd-22, so the drop falls harder on images carrying
several novel boxes than if spread evenly; $25.5$ and $37.3\%$ of the images with a novel object covered beforehand lose
at least one (Appendix~\ref{sec:app:imageweighted}).

Every number here is within-architecture, comparing a checkpoint against its own fine-tuned
descendant, which is the only comparison the design supports.

\textbf{An untested hypothesis.} One hypothesis would explain the spread, and we can state it but not test it here. If most
of the IoU~0.5 fall is a selection loss (\S\ref{sec:proxyvalidation}), then an architecture whose
budget discards nothing has no selection component to lose, and RF-DETR is exactly that
architecture: $K/N = 1.00$ against YOLOX's $0.04$ (Table~\ref{tab:candidates}). The prediction is
that scoring another architecture over its whole candidate set should bring its fall down to
RF-DETR's. On YOLOX that is measured, and it does: on the three seeds scored both ways, $10.50$ and $17.82$
points at $K = 300$ on tabletop-22 and cardd-22 become $0.82$ and $2.36$ over all $8400$
candidates, below RF-DETR's $1.29$ and $5.86$ on the same domains. Faster~R-CNN's budget
sweep is a second and weaker instance, reaching the same prediction by relaxing $K$ instead of
enumerating candidates (\S\ref{sec:proxyvalidation}). Part of what reads as an architecture
difference may therefore be a protocol effect, a consequence of holding $K$ fixed across candidate sets that differ by a factor
of $28$, from RF-DETR's $300$ to YOLOX's $8400$. Neither instance varies $K/N$ with the
architecture held fixed, so this falls short of a test and is stated so that it can be attacked.
Four architectures cannot establish that the effect is architecture-independent, and we do not
claim it.

Every fine-tuned row in the table would have passed inspection from inside its deployment domain.
OWLv2's in-domain mAP rose from $63.79$ to $79.77$ while its IoU~0.7 coverage halved, and on
tabletop-22 Faster~R-CNN reached $84.84 \pm 0.79$ mAP, RF-DETR $91.00$ and YOLOX $80.73$ while
giving up $17.67$, $5.12$ and $12.87$ points at IoU~0.7. A practitioner reading only the in-domain number
would have shipped the last row of each block.

\subsection{Reproduction on a second image distribution}
\label{sec:voc}

Every number above is measured on COCO \texttt{val2017}. Repeating the measurement on Pascal VOC~\citep{everingham2010pascal}
and on OpenImages V4~\citep{kuznetsova2020open}, the latter collected and annotated with no COCO involvement, gives a loss on every cell: all $60$ seed-level drops on VOC and all $36$ on OpenImages are positive,
and on OpenImages Faster~R-CNN loses \emph{more} on cardd-22 than on COCO, $40.17$ against $30.08$
at IoU~0.7. The architecture-dependent domain ordering of
\S\ref{sec:depthdistance} reproduces exactly on images it was not derived from. The tables, the
category map and the two structural limits of both evaluations are in
Appendix~\ref{sec:app:seconddist}.

\subsection{Reproduction under the metric the open-world literature uses}
\label{sec:aratk}

$C_\tau$ reports two thresholds, and a reader is entitled to ask whether the result is a property
of that pair. Standard AR@K is top-$K$ recall meaned over IoU~$0.5{:}0.05{:}0.95$. Recomputed on
the same population it gives a loss on every cell we ran it on: $-11.77$ and $-21.18$ points for
Faster~R-CNN on the two domains, $-12.95$ and $-21.26$ for YOLOX. Each sits between that cell's
$\Delta C_{0.5}$ and $\Delta C_{0.7}$ in level and in drop. The farther domain costs more under
AR@K exactly as it does under $C_\tau$. These four cells are single checkpoints and not the seed
means of Table~\ref{tab:collapse}, because AR@K needs its own forward pass. They are therefore not
comparable to that table row by row (\S\ref{sec:app:ark}). The protocol also repeats on an independently
released YOLOX checkpoint of a different size, which reproduces every ordering at smaller
magnitudes (Appendix~\ref{sec:app:secondckpt}). It repeats on Grounding DINO too, whose rankings are
prompt-conditional by construction and which is therefore read separately rather than counted
as a fifth confirmation (Appendix~\ref{sec:app:gdinodetail}).

\section{What does not explain the drop, and the one split that does}
\label{sec:mechanism}

Held-out coverage falls under adaptation, and some property of the regions or of the model decides which of them are lost. One component of that decision is identified here and comes first: on the architecture whose full candidate set can be enumerated, the fall at IoU~0.5 is
overwhelmingly a ranking loss and the drop at IoU~0.7 is about half a placement loss. Two
measurements sharing no code agree on that split.

The rest is elimination, and what follows is organised as one. Four explanations a reader
would reach for do not survive measurement. The drop does not land on categories the deployment
vocabulary omits, over $333$ categories. Suppressing unmatched regions as negatives is a pathway
carrying $4.6$ to $15.7\%$. The deployment domain's distance does not set the magnitude, since no
ordering of domains survives a change of architecture. Freeze depth yields no rule
(\S\ref{sec:ladder}). A fifth, geometric,
survives \emph{as a direction and not as a quantity}: adaptation moves box coordinates, and a region is lost when that motion carries its overlap
across the threshold. The sign of that relationship holds in every condition measured. Its
coefficient is a property of the fitting window, not of the detector.

The majority of the drop at the strict threshold we leave unaccounted for.

\subsection{Scoring every candidate: what survives when nothing is ranked away}
\label{sec:proxyvalidation}

$C_\tau$ can fall four ways: the model stops emitting a region near the object, it emits one and
ranks it below $K$, it emits and ranks one and places it worse, or post-processing removes it.
Only the first and third are losses of anything a reader would call localisation, and the distinction has so far been handled by wording, although it can be measured.

On an anchor-based detector it can be measured. YOLOX decodes a box at each of $8400$ anchors, so
the set of every region the model can produce is one we can score directly. Scoring $C_\tau$ over
all of them removes selection from the metric: a box counts if the model places a region on it
anywhere in its output, at any score.

A component that is mostly \emph{which} candidates are selected, on candidates the model still
emits, is a ranking problem and not a representation one. Ranking under distribution shift is
studied in its own right: detector confidence is known to miscalibrate when the test distribution
moves~\citep{munir2022towards,kuppers2020multivariate}. We do not test that account
here, and note only that it predicts what the threshold split shows, a loss recoverable by
reordering at IoU~$0.5$ and not at $0.7$.%

\textbf{A second architecture agrees, through a different measurement.} The budget sweep bounds the
same quantity without any candidate correspondence: raising $K$ from $300$ to $3000$ on Faster~R-CNN
closes $72\%$ of the tabletop-22 gap and $69\%$ of the cardd-22 gap at IoU~0.5, against $35\%$ and
$27\%$ at IoU~0.7 (\S\ref{sec:budget}). A gap that a larger budget closes was a ranking loss, so
those figures are a lower bound on selection's share: a budget sweep cannot separate ranking from
post-processing, and both sit on the same side of the split. That bound and the
$87$--$92\%$ YOLOX recovers once $K$ is relaxed to every candidate are two architectures and two
methods sharing no machinery, and they agree that most of the IoU~0.5 fall is selection. At
IoU~0.7 both fall to roughly half, which is the same reversal.

Equation~\eqref{eq:bestiou} maximises per ground-truth box, so one region may be credited to several neighbours; capping each region to one box, greedily
by IoU, moves every entry by at most $0.06$ points. Scoring with no area floor at all,
$31{,}367$ boxes instead of $21{,}861$, moves the IoU~$0.7$ loss by under a point on this
architecture, $-14.60$ against $-13.86$ on tabletop-22 and $-21.88$ against $-22.22$ on cardd-22,
seed $0$. At IoU~$0.5$ the floor is not neutral, and it works against the headline. The same two cells go
from $10.40$ to $13.53$ and from $17.13$ to $19.88$ once it is removed. Over the four cells
measured that way, removing it raises the IoU~$0.5$ loss by $16$ to $113\%$
(Appendix~\ref{sec:app:area}). Neither statement travels: the $48.8$ and $50.9$ point falls below
$256\,\text{px}^2$ are Faster~R-CNN and OWLv2 (\S\ref{sec:sizebands}).

\subsection{Holding the candidate set fixed: how much is selection, how much is geometry}
\label{sec:decompose}

Two architectures carry an anchor index that names the same rectangle before and after
fine-tuning, so the crossing runs on both, at five seeds each: YOLOX's grid of $8400$, and Faster~R-CNN's $242{,}991$
anchors over the FPN. All four baseline cells reproduce the reported coverage of the same
checkpoints exactly (Appendix~\ref{sec:app:decompose}).

\textbf{Component split at IoU~$0.5$.} Pairs give tabletop-22 first. Selection is the larger component on both, and by more on Faster~R-CNN: $63.2$ and $60.2\%$ of the total on YOLOX, $81.5$ and $74.8\%$ on Faster~R-CNN, each a mean over
five seeds with a standard deviation under $1.5$ points. Which anchors the fine-tuned
model puts in its top $300$ accounts for more of the drop than where it places the boxes. That
figure is a main effect at fixed $K$. It is not the $87$--$92\%$ of \S\ref{sec:proxyvalidation},
which is what relaxing $K$ to every candidate recovers and so carries the interaction with it.

\textbf{Geometry's share grows sharply with the threshold on both}, from $1.0$ and $8.8\%$ to
$29.8$ and $39.7\%$ on YOLOX, and from $6.9$ and $16.2\%$ to $71.0$ and $84.7\%$ on Faster~R-CNN.
Geometry's $1.0 \pm 0.9\%$ on YOLOX tabletop-22 is the one share indistinguishable from zero, and
the only cell where a seed disagrees with its own mean on the sign. That is
what a threshold account predicts: a coordinate change that leaves a box above $0.5$ can still
carry it below $0.7$.

\textbf{Component split at IoU~$0.7$.} The two architectures disagree about which component is larger, and we report that rather than a single number. On YOLOX selection stays ahead, $51.8$ and $49.2\%$ against
geometry's $29.8$ and $39.7\%$. On Faster~R-CNN geometry overtakes it, $71.0$ and $84.7\%$ against
selection's $57.2$ and $46.0\%$. All five seeds of every cell agree with their own mean on which
component is larger. The claim that survives both architectures is the IoU~$0.5$ one and the
direction of the threshold effect, not a ranking of the two components at the tight threshold.

\textbf{The interaction term.} The term flips sign between the architectures, and the reason is in the construction. On YOLOX it is positive in every cell, $11.1$ to $35.8\%$: a share of the drop
that needs both the anchor set and the coordinates to have moved, so the two components do not
add. On Faster~R-CNN it is positive at IoU~$0.5$, $11.5$ and $9.0\%$, and negative at IoU~$0.7$,
$-28.2$ and $-30.7\%$, where the two components overlap instead. The sign is unanimous
across all five seeds of all eight cells, so it is a property of the architecture and not of a
draw. Selection on Faster~R-CNN is not a
ranking. The RPN selects by objectness and then runs NMS, which reads box coordinates, so the
selection factor carries geometry inside it and the split is not orthogonal by construction. It is
the quantity a downstream ROI head actually receives, which is the reading worth having, and the
negative interaction is what a non-orthogonal factor looks like when it is measured.

This locates the geometric account of \S\ref{sec:threshold} without confirming or refuting it.
Coordinates carry a substantial share of the tight-threshold drop and almost none of the loose one,
so the account describes one regime, although the measurement as a whole outruns it.

What \emph{selection alone} measures needs stating precisely. Taking the fine-tuned model's
anchors and the pretrained model's coordinates partly measures that the fine-tuned model selects
anchors the pretrained model never placed well, which is a joint property of the pair, and ranking alone does not carry it. The large
interaction term is the same fact from the other side. This is
a factorial decomposition with a real interaction, not an orthogonal split. The swap needs the box
a model \emph{would} place at an anchor, selected or not, which every anchor-based decoder supplies at once. It does not need the anchor to appear in both models' top~$300$, and that is why the
$11$--$17\%$ common-anchor share of \S\ref{sec:displacement} does not bound it. RF-DETR and OWLv2
have no index that would support it at all.

\subsection{The loss falls on what the checkpoint already had}
\label{sec:retention}
\begin{figure}[t]
\centering
\includegraphics[width=\columnwidth]{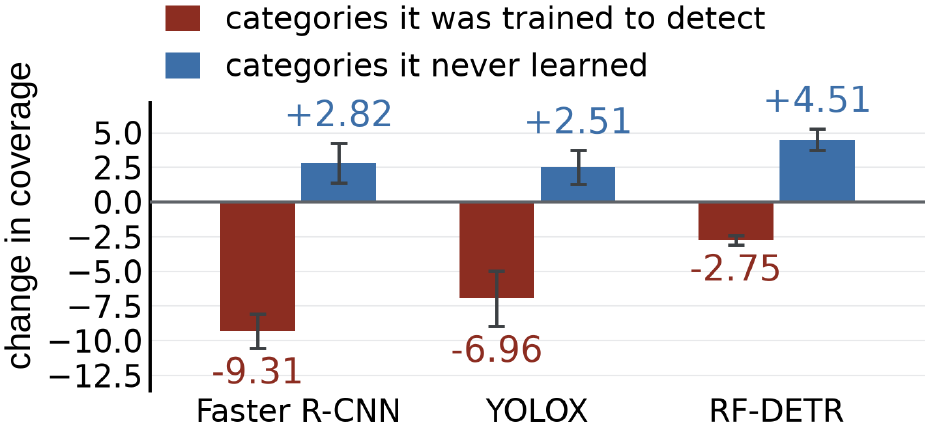}
\caption{\textbf{Adaptation damages what the checkpoint already had, and leaves alone what it never
had.} The $752$ LVIS categories split by whether the pretrained model was trained to detect them.
Both groups sit outside every deployment vocabulary and are read by the same scorer on the same
images, so the split is on pretraining exposure alone. Bars are the mean over five fine-tuned
seeds, whiskers a $95\%$ interval; every bar clears zero and every seed shares its bar's sign
(Table~\ref{tab:retention}).}
\label{fig:retention}
\end{figure}

Every category scored so far was in pretraining, so the measurement is retention. Whether that is a
property of the phenomenon or only of the protocol is answerable from the LVIS runs already
reported: of the $752$ categories they score, $61$ name a COCO-80 class the pretrained model was
trained to detect and $691$ do not. Both groups lie outside every deployment vocabulary, both are
read by the same class-agnostic scorer on the same images, and the split is on pretraining exposure
alone.

\textbf{Which categories lose.} The loss is confined to the categories the model was pretrained on (Figure~\ref{fig:retention}, Table~\ref{tab:retention}). On the $691$ it never learned,
coverage does not fall on any of the
three architectures; it rises, by $2.51$ to $4.51$ points at IoU~$0.5$, while the $61$ it did learn
lose $2.75$ to $9.31$. All six cells are five seeds, all six intervals clear zero, and all five
seeds share their cell's sign. Narrow fine-tuning damages what the checkpoint had and
leaves alone what it never had. That is also why a benchmark of harder categories would show a smaller effect on the same checkpoint.

Three accounts of that pattern were tested and none of them replaces it. A two-rate description
predicts a group's change from its prior coverage alone and gets the sign right in all six cells
and in all thirty seed-level readings, on a split its rates were never fitted to distinguish. It is a conditional-mean description that fails across box scales, and both halves are in
Appendix~\ref{sec:flipmodel}. An indicator for \emph{named by the deployment vocabulary} buys
nothing once prior coverage is conditioned on, on every architecture we could fit, which is a
quantitative exclusion however a null result usually reads
(Appendix~\ref{sec:identity}). Normalised box displacement is positively associated with
overlap loss in all ten conditions, though the coefficient the idealised geometry fixes at $2$ is
reached by none of them (Appendix~\ref{sec:displacement}). What survives all three is the split
above, and the next subsection asks whether the categories it picks out are the same ones across
architectures.

\subsection{Architectures agree on which categories break}
\label{sec:crossarch}
\begin{table}[t]
\centering
\footnotesize
\setlength{\tabcolsep}{3pt}
\caption{\textbf{Different detectors lose the same categories.} Each cell correlates two architectures'
per-category drops over the $333$ LVIS categories with at least ten boxes, at IoU~$0.5$.
\emph{raw} is that correlation; \emph{partial} first removes each architecture's own pretrained
coverage, and \emph{both} removes both. $p$ comes from a $10{,}000$-permutation null that shuffles
category labels within one architecture. At IoU~$0.7$ five of the six cells are larger, the
exception being Faster~R-CNN against YOLOX on cardd-22 (Table~\ref{tab:crossarchfull}).}
\label{tab:crossarch}
\begin{tabular}{@{}p{2.2cm}lrrrr@{}}
\toprule
pair & domain & raw & partial & both & $p$ \\
\midrule
\multirow{2}{2.2cm}{Faster R-CNN\\/ YOLOX} & tabletop-22 & $0.528$ & $0.507$ & $0.594$ & $0.0001$ \\
 & cardd-22 & $0.626$ & $0.581$ & $0.599$ & $0.0001$ \\
\addlinespace
\multirow{2}{2.2cm}{Faster R-CNN\\/ RF-DETR} & tabletop-22 & $0.482$ & $0.323$ & $0.324$ & $0.0001$ \\
 & cardd-22 & $0.536$ & $0.412$ & $0.418$ & $0.0001$ \\
\addlinespace
\multirow{2}{2.2cm}{YOLOX\\/ RF-DETR} & tabletop-22 & $0.319$ & $0.180$ & $0.191$ & $0.0009$ \\
 & cardd-22 & $0.389$ & $0.264$ & $0.270$ & $0.0001$ \\
\bottomrule
\end{tabular}
\end{table}

\textbf{What is new here.} Class-level agreement is not new; the quantity it is shown on is. Both \citet{toneva2019empirical} and \citet{daga2026notall} report it for classification accuracy, on a label the model still has. What is measured here reads no label at all (\S\ref{sec:app:related}).

Naming no property leaves open whether one exists, and the two can be separated
without naming anything. If category identity carries real structure, architectures should agree on
\emph{which} categories lose coverage; if the spread is estimation noise, they should not. The
design is \citeauthor{toneva2019empirical}'s, one level up: they define a forgetting event per
training \emph{example} and establish that a dataset's ``(un)forgettable examples generalize
across neural architectures''~\citep{toneva2019empirical}. The parenthesis is theirs, and it
matters here: the property holds for the examples that break as well as for those that do not. We ask the same question of categories under
adaptation rather than of examples under training, so the test is inherited and what it is applied
to is not. Agreement between models under shift is itself a studied regularity:
\citet{baek2022agreement} report out-of-distribution agreement between any two networks tracking
their in-distribution agreement wherever accuracy is on the line, which is the regularity
\S\ref{sec:invisibility} finds these cells outside. Their agreement is between predictions on the
same inputs and ours is between per-category coverage changes, so the two are one idea at two
granularities and not one measurement.
Table~\ref{tab:crossarch} correlates the per-category change vectors across the three architectures
on the same $333$ categories, removing each architecture's own prior coverage first, since
well-covered categories have more room to fall on any model and would correlate through that alone.
Every one of the twelve cells is positive, with partial $r$ from $0.180$ to $0.581$ and a
permutation $p$ of at most $0.0009$ against a null that shuffles category labels within one
architecture. Repeating the whole comparison on five independently fine-tuned seeds of
all three architectures leaves every one of the twelve cells positive on every seed, at
$0.243$ to $0.606$ with a standard deviation between $0.020$ and $0.069$
(Appendix~\ref{sec:app:crossseeds}). Seed~$0$, which the table above reports, is the weakest draw
for the YOLOX/RF-DETR pair: $0.180$ against a five-seed mean of $0.243$. Faster R-CNN and YOLOX agree most and both agree with RF-DETR less, which is the
ordering RF-DETR shows throughout.

Raising the support floor from
ten boxes per category to a hundred raises every pairwise estimate, from $0.507$, $0.323$ and
$0.180$ to $0.724$, $0.621$ and $0.262$: a correlation attenuated by noise in low-support
categories behaves this way and a spurious one does not. The second pairs one architecture's
tabletop-22 change with another's \emph{cardd-22} change, so that no two vectors come from the same
fine-tune, which should destroy the correlation if a category breaks because of the particular
adaptation. It does not, giving $0.161$, $0.270$ and $0.162$ against same-domain $0.507$, $0.323$
and $0.180$, and on the YOLOX/RF-DETR pair the two are indistinguishable.

Three architectures with three pretraining runs, three proposal mechanisms and three scoring heads
therefore break a shared subset of categories, and for two of the three pairs most of that sharing
survives changing the deployment domain as well. The per-category spread is structure, not noise,
and it belongs to the category, surviving a change of checkpoint and of adaptation, which is also
the strongest evidence here against reading the effect off the individual checkpoints tested
(\S\ref{sec:limitations}). What those categories share we cannot say: partial $r$ near $0.3$ leaves
most of the variance unaccounted for, and elongation fails to supply it.

\subsection{Testing the background-suppression account}
\label{sec:negatives}

The account the open-world proposal literature would give is that a fine-tune teaches the model to
call the held-out objects background: they appear in the training images and no target names them, so the loss pushes every region on them to background. That pathway can be held shut. An
anchor is dropped from the objectness loss when the trainer's own assignment leaves it unmatched
\emph{and} the COCO-pretrained model scores it at $0.5$ or above, the decision boundary of both
objectness branches, fixed before any run and never tuned. Box regression is untouched and nothing
else about the run changes. The rule removes $0.08$ to $0.39\%$ of anchors and, because a
saturating BCE term weights an anchor by its own predicted probability, $37.4$ to $67.0\%$ of the
negative objectness gradient. Two architectures and two domains at five seeds per arm give four
cells, and they are in Appendix~\ref{sec:app:negatives}.

Those two architectures have assignment rules that share nothing, SimOTA against a fixed anchor
grid, and all twenty seed pairs move the same way at IoU~$0.5$ (one-sided sign test,
$p = 9.5\times10^{-7}$), nineteen of twenty at IoU~$0.7$ ($p = 2.0\times10^{-5}$). Shutting the
pathway recovers $4.6$ to $15.7\%$ of the drop at IoU~$0.5$, a mean of
$1.24$ points, without costing in-domain accuracy. Between $84$ and $95\%$ of the drop
survives with the mechanism the open-world literature treats as decisive held shut. That does not make the remainder coordinate drift by elimination, and the bound is measured at
the deepest rung on two of the four architectures. What it rules out is the form of that account the
open-world literature states, at the depth where the drops are largest. \S\ref{sec:ignorearm} reaches the same conclusion on the recognition side by a
different route.

The pathway shut above is one of two. Faster~R-CNN's ROI head trains its own background class on
unmatched proposals by the same logic the RPN's objectness does, and nothing above touches it.
The same rule extends to the ROI head: a background-labelled proposal overlapping a
high-confidence pretrained anchor leaves the classifier and the box-regression pool as well. That
roughly doubles what is recovered, to $19.5\%$ and $23.0\%$ of the drop on cardd-22 and tabletop-22
at IoU~$0.5$, from $10.3\%$ and $12.8\%$ with the RPN alone. The two pathways are
therefore not redundant; an account that closes only one of them was leaving a comparable-sized
recovery on the table. Even at this deeper intervention, three quarters of the drop survives
(Appendix~\ref{sec:app:roihead}).

\section{Freeze depth is a control, not a recipe}
\label{sec:ladder}

Freezing part of the network is the standard advice for adapting a checkpoint without damaging what
it already has, and it is advice about depth: train less of the model and lose less of it. Seven
freeze ladders over four architectures and two deployment domains say the depth-to-damage ordering
belongs to the architecture rather than to depth (Table~\ref{tab:ladders}).

\subsection{The ordering is architecture-dependent}
\label{sec:twoladders}
\begin{table*}[t]
\centering
\small
\setlength{\tabcolsep}{4.5pt}
\caption{\textbf{Freezing more of the network does not reliably protect it}: for three of these seven ladders the deepest stage was not the most damaging. Cells are the change in coverage at IoU~$0.7$ against that
architecture's \emph{own} pretrained model, so rows are not comparable with one another.
\emph{rungs read} leaves out stages whose trained parameters the objectness score cannot see, which
must measure $-0.00$ (\S\ref{sec:invariance}). $\dagger$ marks a rung below $10$ in-domain mAP (\S\ref{sec:control});
those runs are kept here, and dropping them moves two rungs. Five seeds throughout except OWLv2 below \texttt{full} and Faster~R-CNN's interior rungs. Full ladders, and the two rungs: \S\ref{sec:app:ladders}.}
\label{tab:ladders}
\begin{tabular}{llcrrrc}
\toprule
architecture & domain & rungs read & shallowest & worst rung & \texttt{full} & worst is deepest \\
\midrule
\multirow{2}{*}{Faster R-CNN}
 & tabletop-22 & 5 of 7 & $-6.94$ (\texttt{+RPN}) & $-17.67$ (\texttt{full}) & $-17.67$ & yes \\
 & cardd-22 & 5 of 7 & $-36.98$ (\texttt{+RPN}) & $-36.98$ (\texttt{+RPN}) & $-30.08$ & \textbf{no} \\
\midrule
OWLv2 & tabletop-22 & 6 of 6 & $-0.70$ (\texttt{heads}) & $-52.24$ (\texttt{full}) & $-52.24$ & yes \\
\midrule
\multirow{2}{*}{RF-DETR}
 & tabletop-22 & 6 of 6 & $-2.01$ (\texttt{heads}) & $-10.43$ (\texttt{dec3}) & $-5.12$ & \textbf{no} \\
 & cardd-22 & 6 of 6 & $-6.94$ (\texttt{heads})$^\dagger$ & $-43.60$ (\texttt{dec1})$^\dagger$ & $-12.00$ & \textbf{no} \\
\midrule
\multirow{2}{*}{YOLOX}
 & tabletop-22 & 5 of 5 & $-0.04$ (\texttt{preds}) & $-12.87$ (\texttt{full}) & $-12.87$ & yes \\
 & cardd-22 & 5 of 5 & $-8.06$ (\texttt{preds})$^\dagger$ & $-23.79$ (\texttt{full}) & $-23.79$ & yes \\
\bottomrule
\end{tabular}
\end{table*}

On OWLv2 and on YOLOX the ordering is the expected one: every deeper rung costs more coverage than
the one below it and \texttt{full} is the worst. On RF-DETR it inverts. The deepest rung is close to
the cheapest on both domains, $-5.12$ against a worst of $-10.43$ on tabletop-22 and $-12.00$
against $-43.60$ on cardd-22, with the damage peaking in the middle of the ladder. Faster~R-CNN
gives a third pattern, monotone on tabletop-22 and inverted on cardd-22, where the shallowest rung
the measurement can read is also the most damaging one. Three of the seven ladders put their worst
rung somewhere other than the deepest, and the four architectures cannot be reordered into
agreement. Every rung carrying one of those three inversions has five seeds: RF-DETR's
whole ladder does, and Faster~R-CNN's \texttt{+RPN} and \texttt{full} rungs are its two replicated
ones. The thinly seeded rungs are four of OWLv2's six, at one or three seeds, and Faster~R-CNN's interior,
and neither carries an inversion, so the disagreement does not rest on the single runs.

Faster~R-CNN carries one structural caveat: its two shallowest rungs train only parameters the
objectness score does not read, so their coverage cannot move, and those cells are marked
\emph{not measurable} (\S\ref{sec:invariance}). The first rung we can measure already costs
$6.94 \pm 0.11$ points, so whether a genuinely free region exists there is a question this
measurement cannot answer.

\textbf{The two estimands.} All-attempts is the primary estimand below, and admissible-only is a conditional analysis reported beside it. The \emph{all-attempts} view keeps every run a depth produced, failures included, and answers the
question a practitioner faces: choose a rung before knowing whether the adaptation will succeed.
The \emph{admissible} view drops the five cells below $10$~mAP (\S\ref{sec:control}) and answers a
question about depth among successful adaptations. It conditions on a post-treatment variable and
is a conditional analysis, not an estimate of what choosing a depth does. Nothing here is
randomised. \emph{Admissible} describes what the estimand retains; the clinical term would imply a
randomised assignment this design does not have. We lead with
all-attempts everywhere and give the admissible figure beside it. The two agree except where
noted, and the filter is conservative where they differ: dropping those five cells makes the
mean ladder loss \emph{smaller}, $-6.71$ against $-7.46$ at IoU~$0.5$ and $-12.14$ against $-14.14$
at IoU~$0.7$, so the filtered view is the milder one. Table~\ref{tab:collapse} is identical under
both, since every \texttt{full}-rung cell is admissible on all four architectures
(Appendix~\ref{sec:app:itt}).

Depth recovers its expected sign only once in-domain mAP joins the regression, where $R^2$ rises
from $0.05$ to $0.80$ on cardd-22 and the two predictors take opposite signs. A deeper rung
perturbs more and fits the deployment domain better, fitting better is protective, and the two
effects can cancel or invert. Every depth result below therefore holds quality of fit fixed, although a practitioner choosing a
rung holds nothing fixed. Freeze depth is
one control variable whose effect is architecture-dependent, and no generic freeze ladder
survives the seven ladders measured here.

\subsection{A measurement procedure in place of a freeze-depth rule}
\label{sec:procedure}

We recommend a measurement instead. For a candidate deployment domain, six steps.
\textbf{(i)}~Assemble an evaluation set of categories the deployment vocabulary does not name;
COCO \texttt{val2017} supplies one free. \textbf{(ii)}~Fine-tune at each freeze stage, about half
an hour per stage on one RTX~A6000 at this data scale, recording in-domain mAP and held-out
coverage. \textbf{(iii)}~Discard as uninformative any stage that trains neither the parameters the
objectness score reads nor those its boxes come from. \textbf{(iv)}~Drop every stage whose
in-domain accuracy says the fit failed, before reading its coverage; here, the five RF-DETR
and YOLOX cells below $10$~mAP. \textbf{(v)}~Keep the Pareto front of the survivors.
\textbf{(vi)}~Ship from that front, with the trade between axes made by whoever owns the
deployment. On OWLv2 the front collapses to one very shallow stage; on RF-DETR
it contains the full fine-tune, and the ladder's answer is to train everything. A cheap stopping
point may simply not exist, and in the settings studied here that cannot be inferred from
in-domain accuracy alone.

The procedure is stated on the all-attempts estimand, which is the one a practitioner faces:
the stages are run before anyone knows which will succeed, and the discard step belongs to the
procedure itself. Every depth ordering above leads with the all-attempts figure and gives the admissible one beside it. The admissible view conditions on a post-treatment variable and describes
depth among adaptations that already worked, which is a different question from the one the
recommendation answers.

\section{A repair that costs no training run}
\label{sec:mitigations}
\begin{figure}[t]
\centering
\includegraphics[width=\columnwidth]{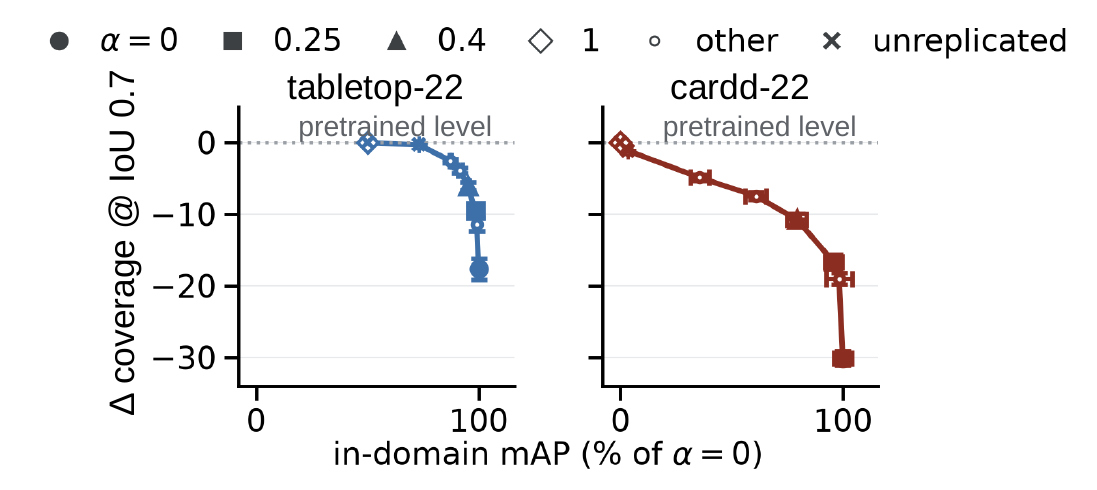}
\caption{\textbf{What mixing the pretrained weights back in buys, and what it costs}, Faster~R-CNN. Each
point is one mixing fraction $\alpha$, running from the fine-tuned model at $\alpha=0$ to the
pretrained one at $\alpha=1$; four fractions carry a marker of their own, keyed above, and the rest are open dots. Up is coverage recovered, left is in-domain accuracy given up, so a curve that rises steeply is buying cheaply. Bars are one standard deviation on each axis. \textbf{The fractions are not measured on a common set of seeds}: five carry $\alpha \in \{0, 0.25, 0.5\}$, three carry $\{0.2, 0.4, 0.6\}$, and the two nearest the pretrained model are single evaluations, marked by a cross. The two domains pay very different prices: tabletop-22 turns almost vertically, cardd-22 trades the two roughly in proportion.}
\label{fig:wiseft}
\end{figure}

Partial freezing fails as a repair. On cardd-22 it is worse than training everything: freezing the
backbone and adapting the RPN costs $36.98 \pm 2.10$ points against $30.08 \pm 0.99$ for the full
fine-tune, non-overlapping over five runs each. A plausible reading is that an RPN forced to adapt
alone, against features that cannot move with it, distorts more than one adapting alongside them,
and we have not tested it.

Mixing the pretrained weights back in costs no training at all. Let
$\theta_\alpha = \alpha\,\theta_{\text{pre}} + (1-\alpha)\,\theta_{\text{ft}}$ and sweep $\alpha$.
What $\theta$ ranges over is the choice that matters here, and three readings of it are distinct.
\emph{Parameter interpolation} mixes the trainable tensors alone and is what the term usually
denotes~\citep{wortsman2022robust,ilharco2022patching}. \emph{State interpolation}, which is what
we report, mixes every float tensor that has a same-shape pretrained counterpart, and that
\emph{includes the normalisation running statistics}. \emph{Buffer restoration} keeps the
fine-tuned parameters and replaces the running statistics with the pretrained ones outright.
Tensors with no same-shape counterpart, the class-predictor weights whose shape changed with the
deployment vocabulary, are carried from the fine-tuned model under all three. On YOLOX that is $382$ tensors mixed against $80$ carried over, with the $148$ BatchNorm running
statistics all on the mixed side. Interpolating pretrained and fine-tuned weights is an inherited repair (\S\ref{sec:related}).
What is new here is narrower: that it recovers \emph{held-out proposal coverage}, and that on YOLOX, the one architecture where the two were separated, the normalisation buffers carry a third of that recovery on cardd-22 and over half on tabletop-22. A third new thing is that the fraction can be selected from a stated in-domain budget, with nothing tuned. The intervention is state interpolation, and the two variants are not the same repair. The weight-averaging literature names both. \citet{izmailov2018swa} recompute the statistics after averaging, and PyTorch ships that as \texttt{update\_bn} beside an \texttt{AveragedModel} flag that averages the buffers instead. Both are therefore available to anyone who averages weights, and what neither literature reports is what the choice between them costs. Re-estimation, the arm that fails below, is therefore that literature's own prescription as well as the adaptation literature's, and two lines of work are contradicted rather than one.
Mixing only \texttt{nn.Parameter} tensors and holding every buffer at its fine-tuned value
attributes $34.9$ to $35.7\%$ of the recovery on cardd-22 and $53.6$ to $55.9\%$ on tabletop-22 to the buffers alone, over five seeds, every seed positive in all four cells (Appendix~\ref{sec:app:paramonly}). The two variants agree exactly at $\alpha = 0$, which checks the split.

\textbf{The re-estimation arm is AdaBN~\citep{li2016adabn}, and on this quantity it is the arm
that fails.} Recomputing a network's normalisation statistics on target data is the standing recommendation of the domain-adaptation literature, across four lines of it (\S\ref{sec:related}). Run here
on the $185$ deployment training images, on top of parameter-only interpolation and with no
training, it recovers $27.8$ to $34.6\%$ of the buffer gap on cardd-22 and is negative on tabletop-22, on every one of five seeds (Table~\ref{tab:bnrecal}). What the buffers carry is therefore not staleness, and the direction that helps is the opposite one: interpolating \emph{toward the pretrained} statistics rather than re-estimating on the target. The arm also fails in the way this paper's own thesis predicts. Paired within seed against the arm we ship, re-estimation is indistinguishable in domain, $+0.08 \pm 0.37$ mAP on tabletop-22 and $-0.44 \pm 0.47$ on cardd-22, both intervals covering zero. Held-out coverage falls by $3.48 \pm 0.56$ and $2.24 \pm 0.38$ points at IoU~$0.5$ with every interval clear of zero. The standard move is free on the axis it is judged on and costly on the axis \S\ref{sec:objectness} introduces. The reading that survives is that these statistics carry something specific to the pretraining distribution, a caveat \S\ref{sec:related} records both literatures already flagging.

Swapping the pretrained buffers in whole recovers most
of the coverage and costs $14.5$~mAP in domain. Interpolation is the only one of the four that
reaches the pretrained buffers' coverage at the fine-tuned buffers' accuracy
(Appendix~\ref{sec:app:bnrecal}). On tabletop-22 the normalisation statistics carry
more of the recovery than the weights do. Parameter-only interpolation still recovers a
substantial share, $+6.28$ and $+2.74$ points at IoU~$0.5$, so the conventional form of the intervention works. The variant reported here is stronger, and mixing the buffers costs no in-domain accuracy, since it also scores $0.95$ and $0.80$ mAP higher at $\alpha = 0.25$
(Appendix~\ref{sec:app:paramonly}). A practitioner implementing parameter-only interpolation from
the weight-averaging literature should expect a weaker result than the numbers below.
That the straight line between the two checkpoints is usable at all is a fact about the
landscape, and it is established. A model trained from pretrained weights ``stays in the same basin in the loss landscape''~\citep{neyshabur2020what}, so the segment between a pretrained checkpoint and its own descendant carries no barrier to cross. The notion of a linearly connected pair is \citet{frankle2020linear}'s, and \citet{zhou2024crosstask} find features interpolating along such paths, on a finetune-to-finetune route where ours runs pretrained-to-finetuned. What that literature does not supply is why the two domains buy coverage back at such different prices. One reading is available and we leave it a conjecture: the shape of the curve would track how far the fine-tune travelled inside the basin,
and the knee would sit where the path leaves the region the pretrained normalisation statistics
still describe. No experiment here separates that from the alternatives, and the buffer
decomposition above is the nearest thing to evidence for it.

The curve has a knee near $\alpha = 0.25$ on every architecture and domain swept
(Table~\ref{tab:archwiseft}, Figure~\ref{fig:wiseft}; Appendix~\ref{sec:app:frcnnsweep} gives
the Faster~R-CNN sweep in full). Past the knee the in-domain column collapses: YOLOX tabletop-22
falls from $79.98$ at $\alpha = 0.25$ to $66.61$ at $0.5$ and $24.18$ at $0.75$, and RF-DETR
cardd-22 from $30.08$ to $12.01$ to $0.15$. The operating point is doing real work in
that sentence, which is why the curve is reported in full.

The cost bounded throughout is read off each cell's own curve at $\alpha = 0.25$, largest of the six at $2.47$ points (Table~\ref{tab:alphaloo}).

\textbf{Leave-one-cell-out selection.} $\alpha = 0.25$ survives being chosen without seeing the cell it is used on. We read it off these curves, so we test it the way a selected constant should be tested. Six
architecture-by-domain cells carry a full curve. Leaving one out and choosing $\alpha$ on the other five (maximising mean coverage gain subject
to a mean cost of one mAP point) returns $0.25$ to $0.27$ on all six folds, and exactly $0.25$
on all six when the search is restricted to measured alphas. Coverage gain on
the held-out cell is positive every time, and within $0.25$ points of the best $\alpha$ choosable
in hindsight under the same budget.

\textbf{Sensitivity to the budget.} $0.25$ is what a one-point budget returns, and a different budget returns a different number. Repeating the same leave-one-cell-out selection at five budgets moves the mean selected
$\alpha$ from $0.152$ at a quarter-point budget to $0.357$ at five points. At every budget the six
folds still agree with each other closely, within $0.05$ of $\alpha$ at four of the five budgets,
so what transfers across architectures and domains is the selection rule, although the constant it returns does not. A practitioner with a different tolerance should read their own value off
Table~\ref{tab:alphabudget} instead of adopting $0.25$.

\textbf{Repair at inference.} Reordering the same candidates recovers part of it without touching a weight. A strictly monotone recalibration cannot move coverage at all, and what remains is bounded three ways;
Appendix~\ref{sec:rescore} states the bounds and prices the one reordering that is free
(Table~\ref{tab:alternatives}, last row).

\section{Limitations}
\label{sec:limitations}

\textbf{What the proxy's restrictions cost.} The area filter is the restriction most likely to
shape a result about box geometry, and removing it raises the IoU~0.5 loss in all four seed-$0$
cells measured, by $16$ to $113\%$, and moves the IoU~0.7 loss by under a point
(\S\ref{sec:proxyvalidation}), so the filtered numbers are the conservative ones. It also creates the one reading the filter is responsible for. RF-DETR's full fine-tune looks nearly free above
$1024\,\text{px}^2$, at $1.29$ points over five seeds, and the seed-$0$ checkpoint that loses $1.73$
there loses $13.10$ below $256$, so \emph{RF-DETR degrades
little} holds for the boxes we measure. Below $256\,\text{px}^2$ it fails. The $21{,}861$ boxes
also come from only $4877$ images, so they share image-level structure: resampling by image widens
a single coverage interval by about half, from $[81.17, 82.20]$ to $[80.88, 82.48]$. That is
negligible against the $5.12$ to $63.35$ point effects, although it is comparable to the gap
between adjacent rungs.
The novel/control split is fixed by one domain's vocabulary, so the control group is
in-vocabulary for tabletop-22 and merely a same-boxes comparison for cardd-22. $K = 300$ is the
largest budget all five detectors, Grounding DINO included, can meet (Appendix~\ref{sec:app:area},
\S\ref{sec:budget}). Everything here is boxes. Whether mask proposals behave the same way is open
and we have no evidence either way: none of the five detectors predicts masks, so answering it
needs a different set of models rather than a further analysis of these.

\textbf{Scope of the ordering result.} The order of the two losses rests on one architecture:
OWLv2 is the only checkpoint here scorable both ways, and it is measured on one domain. The
ordering holds at all six rungs of its ladder, but the shallowest already shows it, so where the two
\emph{begin} to separate is below anything measured here.
Whether the order holds when adaptation is lengthened rather than deepened is untested, since the
architectures with mid-training checkpoints here have no held-out AP to read.

\textbf{One pretrained checkpoint per architecture.} The longitudinal design compares a checkpoint
with its own descendants, which means every result is conditional on the checkpoint it started from.
Four architectures give four independently pretrained models, on different corpora by different
groups, and the effect appears in all four, so no single one of them accounts for it. We test variation
\emph{within} an architecture once, on YOLOX-M (Appendix~\ref{sec:app:secondckpt}). The direction and the
orderings survive that change of checkpoint; a specific magnitude does not, so the $17.95$-point
cardd-22 drop belongs to \texttt{yolox\_s.pth} and describes YOLOX only through it. One pair of
sizes leaves capacity and the pretraining run confounded. Separating recipe from draw would take a second checkpoint at the \emph{same} size and
recipe, and none of these four architectures has one published.

\textbf{The mechanism is measured on three architectures and exactly on two.} YOLOX has a rung
where the selection is frozen, and Faster~R-CNN's anchor index names the same grid cell in both networks. Selection there keeps
only about an eighth of the pretrained top-300 anchors, which bounds the displacement pairing
and not the selection/geometry crossing, since that crossing reads a box the other model would
place rather than one it selected. RF-DETR is matched by assignment, whose errors are displacement-shaped. Calibrated against a
field of known size, the matcher recovers $41.03\%$ of pairs at RF-DETR's magnitude while
reporting median displacement to within about a tenth, enough for population statistics, though too coarse for
individual regions (Appendix~\ref{sec:app:matcher}). OWLv2 and Grounding DINO are untested. Inside the mechanism the bounds run further. The short side beats the square root of the
area only in the two conditions whose correspondence is exact: YOLOX's \texttt{preds} rung, and
Faster~R-CNN's anchor name on both domains. The balance between translation and rescaling is also architecture-dependent, with rescaling costing more overlap on $79.2\%$ of pairs at YOLOX's pinned
rung against $49.6\%$ on RF-DETR tabletop-22. The negative-label ablation
bounds the competing account at the \texttt{full} rung on two architectures with five seeds per
arm, and says nothing about intermediate rungs. On Faster~R-CNN it is scoped to the RPN, so the ROI
head's background class lies outside it.

\textbf{The conditional-mean baseline fits a mean and misses a variance.} The independence it was
derived from is falsified: categories move together, and observed dispersion runs from $7.34$ to
$9.23$ where independent flips predict one. Letting each
category's rate vary around the model's mean, with a single shared concentration, absorbs that and
supplies a usable variance estimate, though it names the heterogeneity without explaining it. The one
category property we found that predicts the residual is elongation, and it accounts for under
$8\%$ of that residual's variance. What decides survival at the aggregate level remains largely
unaccounted for. $q_\uparrow$ in particular is extrapolated to $X = 0$, and no run observes it
directly (Appendix~\ref{sec:app:bootstrap}). It is also a statement about aggregates over
categories and not about any partition of the same boxes, which \S\ref{sec:retention} states with
the size-band measurement that establishes it.

\textbf{Three domains cannot isolate distance, and the seeds are uneven.} cardd-22 is further
from the pretraining distribution than tabletop-22 by both measures we took and loses more at every rung. It is also harder in absolute terms, $22.65$~mAP against $84.84$ on Faster~R-CNN, and differs in annotation style, class overlap, object scale, camera and construction,
all moving together. bccd-3 breaks that pairing: it is furthest of the three and middling for difficulty, so difficulty cannot be what orders the drop. It substitutes two differences of its own, a
$416 \times 416$ image and a three-class vocabulary. Every added domain trades one confound for
another, and a claim that distance \emph{sets} the magnitude still needs three to five domains
spanning controlled difficulty.

A third deployment condition was built to give a third data point and yielded a precondition
instead.
\textbf{voc-20} matches the other two in shape ($247$ Pascal VOC images, $20$ classes, the same
$185/62$ split) and every run fails the validity control of \S\ref{sec:control}: the pretrained
model scores $74.20$~mAP on voc-20's own validation split against the fine-tune's $55.30$ over
three seeds. Scored on seed $0$ against its own training images, each fine-tune memorises them to
within half a point of the other, $99.21$ on voc-20 against $99.75$ on tabletop-22, so overfitting
fails to separate them. What does separate them is generalisation and baseline. voc-20 falls to
$55.63$ on its validation images, a $43.6$-point gap against tabletop-22's $19.4$, carrying $404$
training boxes over $20$ classes against $1089$ over $22$ across a far wider image distribution.
Its pretrained baseline stands at $74.20$ against tabletop-22's $2.31$, because its vocabulary
is COCO's own.

\textbf{What the measurement licenses.} Only a safety-critical perception stack turns this measurement into a release gate. Elsewhere it stays a diagnostic with no guarantee attached, since a model can pass the proxy and
still fail on the road. The cost is one held-out evaluation pass before adaptation and one after.

\textbf{The precondition.} Two things together produce this: a pretrained model already strong on the deployment vocabulary, and supervision too thin to beat it on held-out images. Neither alone produces it; tabletop-22 overfits as hard and still clears its baseline by seventy-eight points.
Where the conjunction holds, fine-tuning trades nothing: it is simply the wrong move, and the
control catches it before any coverage number is read. No voc-20 number appears in any claim here.
Seed coverage is uneven and Table~\ref{tab:evidence} records it per measurement, from five seeds
across the RF-DETR and YOLOX ladders down to one in the interior of the sweeps; every single-run
reading is labelled qualitative where it is used. Excluding the five cells below $10$~mAP
(\S\ref{sec:control}) is post-treatment selection by a threshold set after seeing the data, which is
why every aggregate is also reported with them in. OWLv2 was never fine-tuned on cardd-22 or measured on LVIS, so the
architecture that loses most in the tabletop-22 comparison, $52.24$ points at IoU~$0.7$, is absent
from the two-domain comparison and from the model. The largest loss anywhere in the study is
larger still, $63.35$ points, and it is YOLOX on bccd-3. Grounding DINO has no class-agnostic score at all, so its every number is prompt-conditional.
Taking all $900$ queries removes the ranking and leaves the conditioning in place. Boxes move by up
to $0.97$ in normalised coordinates between prompts, and the control that supplies its semantic
evidence also moved the ranking-free $K = 900$ number from $86.57$ to $90.89$. Its behaviour is
consistent with a substantial prompt-dependent retrieval component alongside a geometric one that
no measurement inside this architecture can isolate; the split between them is not identified.

\section{Conclusion}

Fine-tuning a detector on a narrow deployment domain is the standard move, and it degrades the
model's top-$K$ coverage of objects that domain's vocabulary does not name. It costs the names
first. On the architecture scorable both ways, adaptation takes $87\%$ of the detection AP over those categories against a fifth of the coverage. The ordering holds at all six
depths of its freeze ladder, in every run (\S\ref{sec:namefirst}). A detection benchmark re-run on
such a checkpoint charges the two to one number and attributes neither. In-domain evaluation does not
reveal the drop either: two checkpoints matched to the same in-domain mAP differ by $11.37$
points of held-out coverage. The loss reproduces over every axis in Table~\ref{tab:claims}'s first row, including an evaluation
set collected and annotated outside the COCO ecosystem, where it does not attenuate and the
architecture-dependent domain ordering reproduces too (\S\ref{sec:budget}, \S\ref{sec:voc},
Appendix~\ref{sec:app:openimages}). Every figure is a paired change within one architecture, so
the magnitudes rank nothing (\S\ref{sec:collapse}).

The two thresholds lose different things, on the one architecture whose candidates can be
enumerated. Scored over all $8400$ of YOLOX's anchors rather than its top $300$, the drop at
IoU~0.5 nearly disappears: the fine-tuned model still emits a region on the object and has
stopped ranking it above $K$. At IoU~0.7 between a third and a half survives the removal of
ranking, which no reordering recovers. Neither figure is claimed for the other three
architectures (\S\ref{sec:proxyvalidation}).

One candidate explanation is that a model stops finding what it was not asked to find. Tested on
$333$ categories, a vocabulary indicator adds no explanatory value once prior coverage is
conditioned on, on any architecture we could fit, and the bound is quantitative, although a null result invites the opposite reading. The
indicator can move a category's expected change by at most a third of one standard deviation. What replaces it makes no reference to a vocabulary. Fine-tuning displaces regions, coverage is a threshold on overlap, and a region is lost when the
displacement carries it across. The geometric identity predicts two things. A tighter threshold
costs more, which holds everywhere we measured. The coefficient is $2$, which the fitting window and the matching rule move by more than architecture does, and which is therefore not
identified here (\S\ref{sec:displacement}).

Freeze depth is one control variable, its effect is architecture-dependent, and it can be
dominated by the opposing effect of target-task fit. We therefore recommend a procedure in place of a ladder
(\S\ref{sec:procedure}). State interpolation needs no training run, and its one parameter is read off a deployment-defined accuracy budget, with nothing tuned. No shorter schedule or trajectory average beat it with an interval clear of zero, over $64$ and
$20$ comparisons on YOLOX. Given a training run and a
slice of the pretraining corpus, replay followed by interpolation beats interpolation alone. The recommendation
therefore splits by budget, and no single method wins outright. Restoring a quarter of
the pretrained state, normalisation buffers included, buys back a large share of
the coverage for under one point of in-domain mAP on four of six architecture-by-domain cells, and up to $2.47$ on the other two. On YOLOX the buffers carry a third of it on cardd-22 and over half on tabletop-22, and the direction is the one the literature on them argues against: re-estimating them on deployment data does worse than nothing on one domain (\S\ref{sec:mitigations}). Training-time
weight anchoring beats it in $22$ of $24$ matched-accuracy comparisons on Faster~R-CNN and loses
all $24$ on YOLOX (\S\ref{sec:anchoring}).

The background-suppression account is measured on both sides and carries neither.
On localisation, fine-tuning with unmatched high-confidence regions ignored instead of pushed to background recovers $4.6$ to $15.7\%$ of the drop at IoU~$0.5$, with all twenty seed pairs moving the same way there and nineteen of twenty at IoU~$0.7$. Closing the same pathway in Faster~R-CNN's ROI
head as well as its RPN reaches $19.5$ to $23.0\%$. On recognition, holding it open for held-out class names recovers none of it
(\S\ref{sec:negatives}, \S\ref{sec:ignorearm}).  What the residual is we cannot say, though it is structured: three architectures with
three separate pretraining runs agree on which categories lose coverage, at partial $r$ up to
$0.581$ after each one's prior coverage is removed (\S\ref{sec:crossarch}).
\backmatter

\bmhead{Declarations}

\textbf{Funding.} This work was supported by the Ministry of Trade, Industry and Resources (MOTIR, Korea) under the Strategic Technology Development Program supervised by the Korea Institute for Advancement of Technology (KIAT) [Grant No. P0028443], under the Industrial Technology Innovation Program (No. RS-2023-00232141, RS-2026-25618509), and by the Institute of Information \& Communications Technology Planning \& Evaluation (IITP) grant funded by the Korean government (MSIT) (RS-2024-00336738).

\textbf{Competing interests.} The authors declare no competing interests.

\textbf{Ethics approval and consent.} Not applicable. The study uses public
benchmark imagery and one image set collected by the authors from a robot arm's
wrist camera in a controlled indoor setting, containing no people and no
personal information.

\textbf{Data and code availability.} Every measurement reported here is
released with the paper as Electronic Supplementary Material, one JSON file per
run, with the scripts that produced them and the two audit programs that check
the manuscript against them, so checking any number needs no GPU, no dataset and
no checkpoint. The three public deployment domains rebuild from their sources
with the scripts provided, the fourth was collected by the authors and
accompanies the release, and the checkpoints are regenerated by the training
scripts there.

\textbf{Author contributions.} T.M.B. designed the study, implemented the measurement, ran every experiment reported here and wrote the manuscript. J.M., Y.K. and J.-H.H. contributed to the study design and to reviewing and editing the manuscript. D.S. supervised the work, acquired funding, and is the corresponding author. All authors read and approved the final manuscript.

\bibliography{references}

\end{document}